\documentclass{article}
\usepackage{iclr2023_conference,times}

\usepackage{amsmath,amsfonts,bm}

\def\eqref#1{equation~\ref{#1}}

\def\1{\bm{1}}

\DeclareMathAlphabet{\mathsfit}{\encodingdefault}{\sfdefault}{m}{sl}
\SetMathAlphabet{\mathsfit}{bold}{\encodingdefault}{\sfdefault}{bx}{n}

\usepackage[colorlinks, linkcolor=red, anchorcolor=red, citecolor=cyan, breaklinks]{hyperref}
\usepackage{url}

\usepackage{times}
\usepackage{epsfig}
\usepackage{amsmath}
\usepackage{amssymb}
\usepackage{wrapfig}
\usepackage{enumitem}
\usepackage{graphicx}
\usepackage{subfiles}
\usepackage{multirow}
\usepackage{booktabs}
\usepackage{subfigure}

\usepackage{color}
\usepackage{colortbl}
\setlist[itemize]{leftmargin=2em}

\usepackage{algorithm}
\usepackage{algpseudocode}

\title{Teaching Yourself:\ Graph Self-Distillation on Neighborhood for Node Classification}

\iclrfinalcopy

\author{
  Lirong Wu, Jun Xia, Haitao Lin, Zhangyang Gao, Zicheng Liu, Guojiang Zhao, and Stan Z. Li \\
  AI Lab, School of Engineering, Westlake University \\
  \texttt{\{wulirong,xiajun,linhaitao,stan.zq.li\}@westlake.edu.cn} \\
}

\begin{document}

\maketitle

\vspace{-1em}
\begin{abstract}
Recent years have witnessed great success in handling graph-related tasks with Graph Neural Networks (GNNs). Despite their great \emph{academic success}, Multi-Layer Perceptrons (MLPs) remain the primary workhorse for practical \emph{industrial applications}. One reason for this academic-industrial gap is the neighborhood-fetching latency incurred by data dependency in GNNs, which make it hard to deploy for latency-sensitive applications that require fast inference. Conversely, without involving any feature aggregation, MLPs have no data dependency and infer much faster than GNNs, but their performance is less competitive. Motivated by these complementary strengths and weaknesses, we propose a \textit{Graph Self-Distillation on Neighborhood} (GSDN) framework to reduce the gap between GNNs and MLPs. Specifically, the GSDN framework is based purely on MLPs, where structural information is only implicitly used as prior to guide knowledge self-distillation between the neighborhood and the target, substituting the explicit neighborhood information propagation as in GNNs. As a result, GSDN enjoys the benefits of graph topology-awareness in training but has no data dependency in inference. Extensive experiments have shown that the performance of vanilla MLPs can be greatly improved with self-distillation, e.g., GSDN improves over stand-alone MLPs by 15.54\% on average and outperforms the state-of-the-art GNNs on six datasets. Regarding inference speed, GSDN infers 75$\times$-89$\times$ faster than existing GNNs and 16$\times$-25$\times$ faster than other inference acceleration methods.
\end{abstract}

\vspace{-1em}
\section{Introduction}
\vspace{-0.5em}
Recently, Graph Neural Networks (GNNs) \citep{wu2020comprehensive,zhou2020graph} have demonstrated their powerful capability to handle graph-related tasks. Despite their great \emph{academic success}, practical deployments of GNNs in the industry are still less popular, and MLPs remain the primary workhorse for practical \emph{industrial applications}. One reason for this academic-industrial gap is the neighborhood-fetching inference latency incurred by data dependency in GNNs \citep{jia2020redundancy}. 

\begin{figure}[!tbp]
    \vspace{-2em}
	\begin{center}
		\subfigure[A two-layer GNN]{\includegraphics[width=0.25\linewidth]{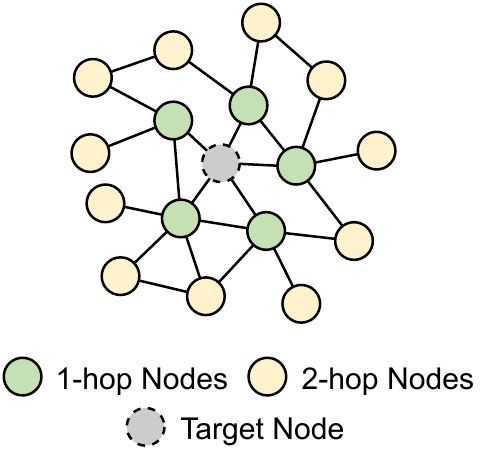} \label{fig:1a}}
		\subfigure[\# Fetched Nodes \textit{vs.} Layers]{\includegraphics[width=0.30\linewidth]{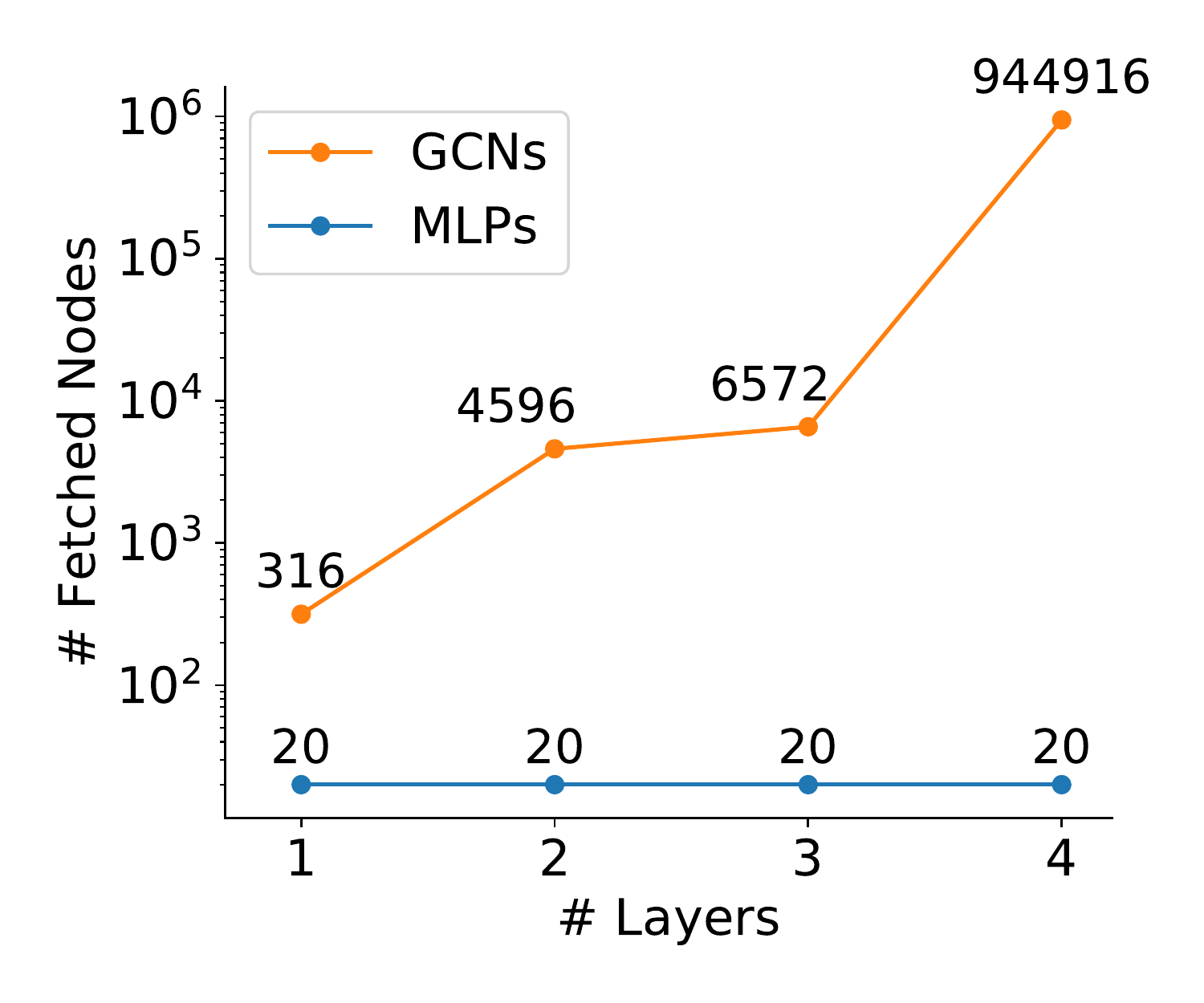} \label{fig:1b}}
		\subfigure[Accuracy \textit{vs.} Inference Time]{\includegraphics[width=0.35\linewidth]{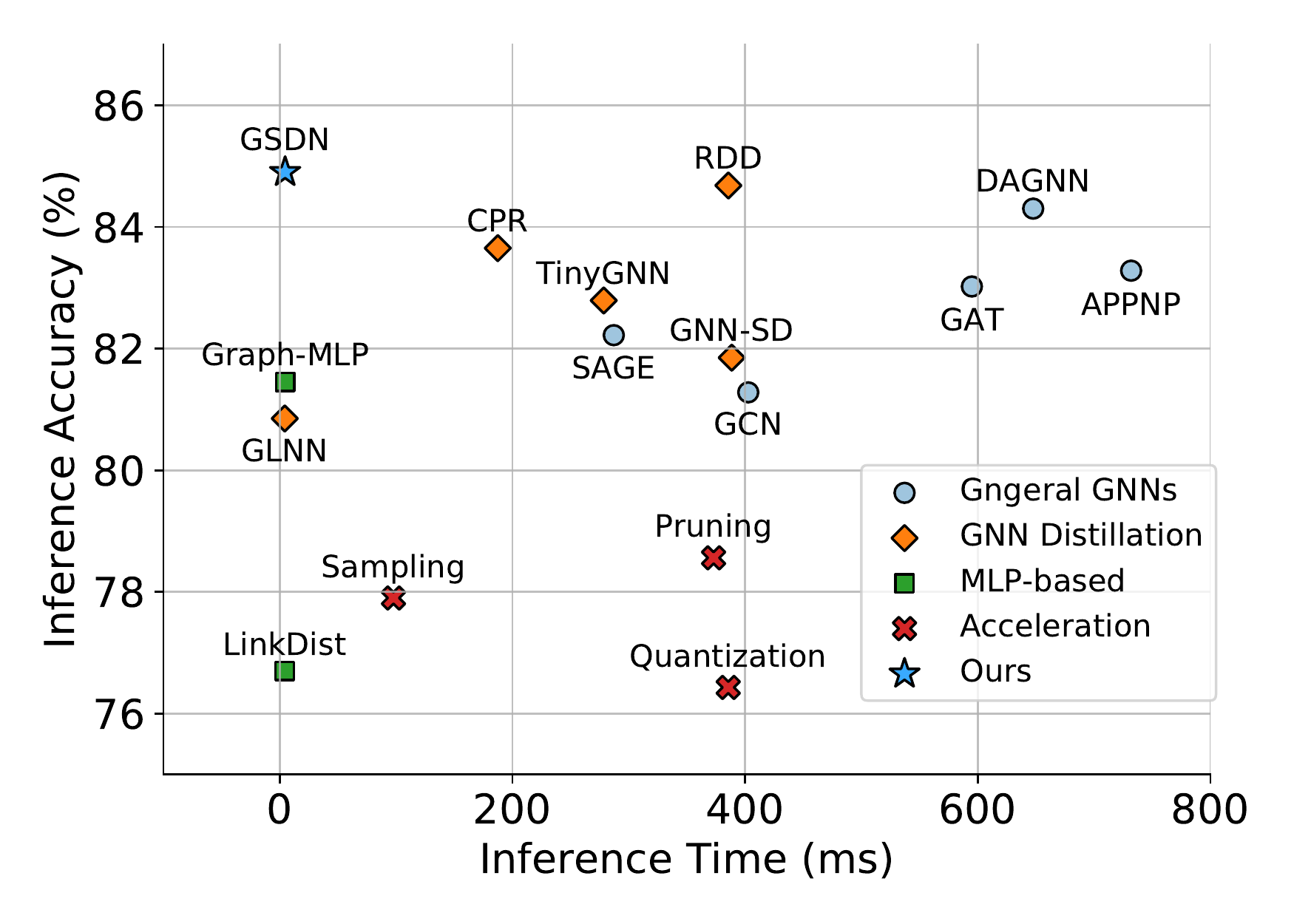} \label{fig:1c}}
	\end{center}
	\vspace{-1em}
	\caption{(a): The one-hop and two-hop neighboring nodes to be fetched by a two-layer GNN. (b): Number of nodes to be fetched for inference with the number of layers, with 20 nodes randomly picked from the \texttt{Coauthor-CS} dataset. (c): Inference accuracy \textit{vs.} inference time on the \texttt{Cora} dataset. If not specifically mentioned, all relevant models adopt GCN as the backbone by default.}
	\vspace{-1em}
	\label{fig:1}
\end{figure}

Most existing GNNs rely on message passing to aggregate neighborhood  features for capturing long-range data dependency between nodes. As a result, neighborhood fetching caused by data dependency is one of the major sources of GNN latency during inference. For example, to infer a single node with a $L$-layer GNN on a graph with average node degree $R$ as shown in Fig.~\ref{fig:1a}, it requires fetching and aggregating $\mathcal{O}(R^L)$ nodes. However, $R$ can be large for real-world graphs, e.g., 19 for Amazon-com dataset, and $L$ is getting deeper for the latest GNN architectures, e.g., $L$=1001 layers for RevGNN-Deep \citep{li2021training}. Moreover, the total latency explodes quickly as $L$ increases, as fetching for successive layers must be done sequentially. Taking Graph Convolutional Networks (GCNs) \citep{kipf2016semi} and MLPs as examples, Fig.~\ref{fig:1b} illustrates the relationship between the layer number $L$ and the number of fetched nodes. Intuitively, it can be seen that the fetch number of GCNs is orders of magnitude more than MLPs and grows exponentially with the layer number. Compared to GNNs, MLPs are free from the data dependency problem and are easier to deploy. However, due to the lack of modeling data dependency, MLPs fail to take full advantage of the graph topological information, which greatly limits their performance on downstream tasks.

Considering both inference accuracy and inference time, general GNNs and MLP-based models are two completely different worlds. As illustrated in Fig.~\ref{fig:1c}, the MLP-based models, such as Graph-MLP \citep{hu2021graph} and LinkDist \citep{luo2021distilling}, are faster in inference but with much poorer performance compared to GNNs. There are two main branches of existing approaches to connect these two worlds. The first branch is inference acceleration, such as Pruning \citep{han2015learning}, Quantization \citep{gupta2015deep}, and Sampling \citep{chen2018fastgcn}, but their improvements in the inference speed are often limited and come at the cost of undesired performance drops. The other branch is graph distillation methods, but they either do not help much in inference acceleration with the data dependency unresolved, such as CPF \citep{yang2021extract}, RDD \citep{zhang2020reliable}, and TinyGNN \citep{yan2020tinygnn}, or they are not competitive with the state-of-the-art GNNs in terms of inference accuracy, such as GLNN \citep{zhang2021graph}. Therefore, \textit{``how can we bridge the two worlds, enjoy the low-latency of MLPs and high-accuracy of GNNs?"} is still a tricky problem.

In this paper, we propose a simple yet effective \textit{Graph Self-Distillation on Neighborhood} (GSDN) framework to reduce the gap between GNNs and MLPs. The proposed GSDN framework is based purely on MLPs, where structural information is only implicitly used as prior to self-distill knowledge between the neighborhood and the target, substituting the explicit information propagation as in GNNs. More importantly, self-distillation can be done in an offline manner during training, and then the trained MLP model can be directly deployed for online inference. In other words, we can shift considerable work from the latency-sensitive inference stage, where time reduction in milliseconds makes a huge difference, to the less latency-insensitive training stage, where time cost in hours is often tolerable. Finally, the resulting model enjoys the benefits of graph topology-awareness in training but reduces time overhead in inference. Extensive experiments have been provided to (1) demonstrate the advantages of GSDN over existing methods in terms of inference accuracy and inference speed and (2) explore \emph{how GSDN can benefit from neighborhood self-distillation}.

\vspace{-1em}
\section{Related Work}
\vspace{-0.5em}
\textbf{Graph Neural Networks.}
GNNs can be mainly divided into two categories, i.e., spectral-based GNNs and spatial-based GNNs. The spectral-based GNNs, such as ChebyNet \citep{defferrard2016convolutional} and GCN \citep{kipf2016semi}, define graph convolution kernels in the spectral domain based on the graph signal processing theory. Instead, the spatial-based GNNs, such as GraphSAGE \citep{hamilton2017inductive} and GAT \citep{kipf2016semi}, directly define updating rules in the spatial space and focus on the design of neighborhood aggregation functions. We refer interested readers to the recent survey \citep{wu2020comprehensive,zhou2020graph} for more GNN architectures. 

Despite their great progress, the above GNNs all share the de facto design that structural information is explicitly used for message passing, which leaves neighborhood fetching still one major source of GNN inference latency. To solve this problem, a large number of \textbf{inference acceleration} methods have been proposed to reduce multiplication and accumulation operations. Specifically for GNNs, pruning \citep{han2015learning} and quantizing \citep{gupta2015deep} GNN parameters or Neighborhood Sampling \citep{chen2018fastgcn} can speed up GNN inference, but they have not completely eliminated the neighborhood-fetching latency, and it often comes at the cost of undesired performance drops.

\textbf{Multi-Layer Perceptron.}
Compared to GNNs, MLPs have no data dependency and are easier to deploy. There have been some attempts to combine pure MLPs with sophisticated techniques, such as contrastive learning and knowledge distillation, which yielded promising results on many graph-related tasks. For example, Graph-MLP \citep{hu2021graph} designs a neighborhood contrastive loss to bridge the gap between GNNs and MLPs by implicitly utilizing the adjacency information. Instead, LinkDist \citep{luo2021distilling} directly distills knowledge from connected node pairs into MLPs without message passing. Despite their great progress, these methods still cannot match the state-of-the-art GNNs in terms of classification performance due to the lack of modeling the graph topology.

\textbf{Graph Knowledge Distillation.}
Several previous works on GNN distillation try to distill knowledge \textit{from large teacher GNNs to smaller student GNNs}, termed as GNN-to-GNN. The student model in RDD \citep{zhang2020reliable} and TinyGNN \citep{yan2020tinygnn} is a GNN with fewer parameters, but not necessarily fewer layers than the teacher GNN, which makes both designs still suffer from the neighborhood-fetching latency. The other branch of graph knowledge distillation is to distill \textit{from large teacher GNNs to lightweight student MLPs}, termed as GNN-to-MLP. For example, CPF \citep{yang2021extract} proposes to distill knowledge from GNNs to a student MLP, but it takes advantage of Label Propagation (LP) \citep{iscen2019label} to improve performance and thus remains heavily data-dependent. Besides, GLNN \citep{zhang2021graph} proposes to distill knowledge directly from GNNs to MLPs, which has a great advantage in inference speed, but its inference accuracy cannot even match the 
vanilla teacher GNNs on a few datasets. Different from the above teacher-student knowledge distillation framework, there is a similar technique known as Knowledge Self-Distillation \citep{zhang2019your}, which regularizes a model by \emph{distilling its own knowledge without other teacher models}. An effective way to achieve this is to regularize the consistency of the predictions from the ``relevant" data, e.g., the label distributions of same-label nodes. For example, GNN-SD \citep{chen2020self} directly distills knowledge across different GNN layers, aiming to solve the over-smoothing problem, with unobvious performance improvement at shallow layers. Based on the smoothness assumption, this paper assumes that neighboring nodes in the graph share similar features and labels, i.e., they can be considered as ``relevant" data. Therefore, GSDN implicitly self-distills knowledge between the neighborhood and the target by regularizing the consistency of their label distributions.

\vspace{-1em}
\section{Preliminaries}
\vspace{-0.5em}
\textbf{Notations and Problem Statement.}
Let $\mathcal{G}=(\mathcal{V}, \mathcal{E})$ denote a graph, where $\mathcal{V}$ is the set of $|\mathcal{V}|=N$ nodes with features $\mathbf{X}=\left[\mathbf{x}_{1}, \mathbf{x}_{2}, \cdots, \mathbf{x}_{N}\right]\in \mathbb{R}^{N \times d}$ and $\mathcal{E}$ is the set of edges between nodes. Each node $v_i \in \mathcal{V}$ is associated with a $d$-dimensional features vector $\mathbf{x}_{i}$. Following the common semi-supervised node classification setting, only a subset of node $\mathcal{V}_L=\{v_1,v_2,\cdots,v_L\}$ with corresponding labels $\mathcal{Y}_L=\{y_1,y_2,\cdots,y_L\}$ are known, and we denote the labeled set as $\mathcal{D}_L=(\mathcal{V}_L,\mathcal{Y}_L)$ and unlabeled set as $\mathcal{D}_U=(\mathcal{V}_U,\mathcal{Y}_U)$, where $\mathcal{V}_U=\mathcal{V} \backslash \mathcal{V}_L$. The task of node classification aims to learn a mapping $\Phi: \mathcal{V}\!\rightarrow\!\mathcal{Y}$ on labeled data $\mathcal{D}_L$, so that it can be used to infer the labels $\mathcal{Y}_U$.

\textbf{Message Propagation.}
A general GNN framework consists of two key computations for each node $v_i$ at every layer: (1) $\operatorname{AGGREGATE}$ operation: aggregating messages from neighborhood $\mathcal{N}_i$; (2) $\operatorname{UPDATE}$ operation: updating node representation from its representation in the previous layer and aggregated messages. Considering a $L$-layer GNN, the formulation of the $l$-th layer is as follows
\begin{equation}
\mathbf{m}_{i}^{(l+1)} = \operatorname{AGGREGATE}^{(l)}\big(\big\{\mathbf{h}_{j}^{(l)}: v_{j} \in \mathcal{N}_i\big\}\big),
\mathbf{h}_{i}^{(l+1)} = \operatorname{UPDATE}^{(l)}\big(\mathbf{h}_{i}^{(l)}, \mathbf{m}_{i}^{(l+1)}\big)
\label{equ:1}
\end{equation}
where $0\leq l \leq L-1$, $\mathbf{h}_{i}^{(l)}$ is the embedding of node $v_i$ in the $l$-th layer, and $\mathbf{h}_{i}^{(0)}\!=\!\mathbf{x}_{i}$ is the input feature. After $L$ message-passing layers, the final node embeddings $\mathbf{h}_{i}^{(L)}$ can be passed through an additional linear inference layer $\mathbf{y}_i=f_\theta(\mathbf{h}_i^{(L)})$ for node classification on the target node $v_i$.

\vspace{-0.5em}
\section{Methodology}
\begin{figure*}[!htbp]
    \vspace{-1em}
	\begin{center}
		\includegraphics[width=1.0\linewidth]{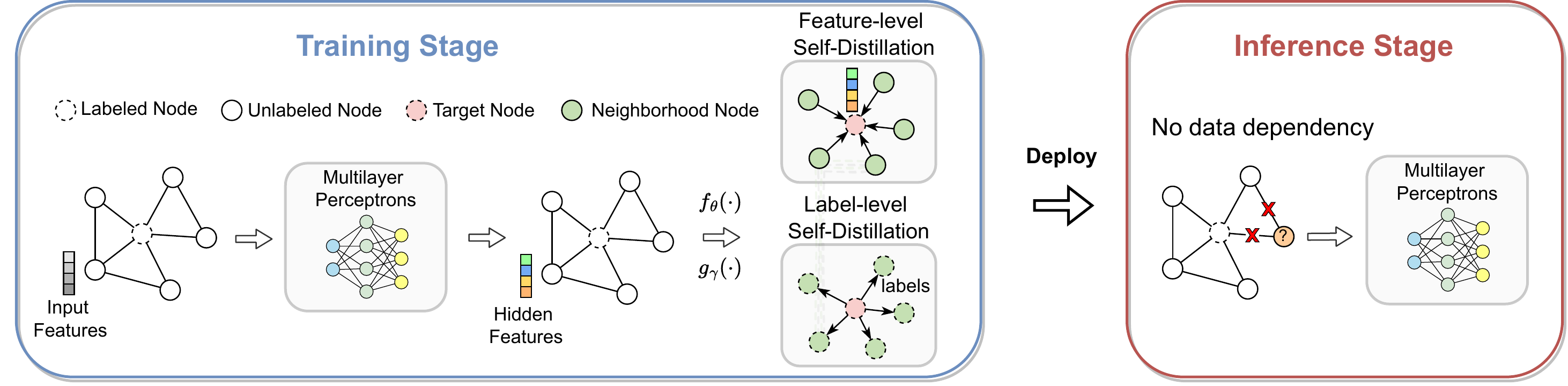}
	\end{center}
	\vspace{-1em}
	\caption{Illustration of the proposed GSDN framework. In the training stage, the MLP and two inference layers $f_\theta(\cdot)$, $g_\gamma(\cdot)$ are jointly trained by neighborhood feature-level and label-level self-distillatio defined in Eq.~(\ref{equ:7}) and Eq.~(\ref{equ:8}). Then, the trained MLP is deployed for inference, where the inference speed of GSDN is much faster than GNNs with data dependency completely removed.}
	\vspace{-1em}
	\label{fig:2}
\end{figure*}

\vspace{-0.5em}
Motivated by the complementary strengths and weaknesses of GNNs and MLPs, we propose a simple but effective \textit{Graph Self-Distillation on Neighborhood} (GSDN) framework to reduce their gaps, with an illustration shown in Fig.~\ref{fig:2}. The proposed GSDN framework enjoys the benefits of graph topology-awareness in training but has no graph dependency in inference. In essence, we need to address the following three important issues: (1) backbone architecture design, how to construct a ``boosted" MLP as backbone architecture; (2) objective function design, how to self-distill knowledge between neighboring nodes and the target node; (3) training and inference, how to solve optimization difficulties in training and conduct predictions with the trained model in inference.

\subsection{Backbone Architecture}
The GSDN framework is based on a pure MLP architecture, with each layer composed of a linear transformation, an activation function, a batch normalization, and a dropout function, defined as:
\begin{equation}
\mathbf{H}^{(l+1)}=\operatorname{Dropout}\big(B N\big(\sigma\big(\mathbf{H}^{(l)} \mathbf{W}^{(l)}\big)\big)\big),\quad \mathbf{H}^{(0)}=\mathbf{X} 
\end{equation}
where $0 \leq l \leq L-1$, $\sigma=\mathrm{ReLu}(\cdot)$ denotes an activation function, $BN(\cdot)$ denotes the batch normalization, and $\operatorname{Dropout}(\cdot)$ is the dropout function. $\mathbf{W}^{(0)} \in \mathbb{R}^{d \times F}$ and $\mathbf{W}^{(l)} \in \mathbb{R}^{F \times F}$ $(1 \leq l \leq L-1)$ are layer-specific weight matrices with the hidden dimension $F$. 

Given a target node $v_i$ and its neighboring nodes $\mathcal{N}_i$, we first feed their features $\mathbf{x}_i$ and $\{\mathbf{x}_j\}_{j\in\mathcal{N}_i}$ into the forked MLP and encode them into hidden representations $\mathbf{h}^{(L)}_i$ and $\{\mathbf{h}^{(L)}_j\}_{j\in\mathcal{N}_i}$. Then, we define two inference layers: $\mathbf{y}_i=f_\theta(\mathbf{h}_i^{(L)}) \in \mathbb{R}^{C}$ and $\mathbf{z}_j=g_\gamma(\mathbf{h}_j^{(L)}) \in \mathbb{R}^{C}$ for label prediction on the target node $v_i$ and neighboring node $j\in\mathcal{N}_i$, where $C$ is the category number. Both inference layers are implemented with linear transformations by default in this paper. Next, we discuss how to implicitly self-distill knowledge between the neighborhood $\mathcal{N}_i$ and the target node $v_i$.

\subsection{Self-Distillation on Neighborhood}
\textbf{Feature-level Self-Distillation.}
The smoothness assumption indicates that neighboring nodes in a graph tend to share similar features and labels, while non-neighboring nodes should be far away. With such an motivation, we perform feature-level self-distillation on neighborhood by regularizing the consistency of the label distributions between the target node and its neighboring nodes. Along with connectivity, disconnectivity between nodes also carries important information that reveals the node dissimilarity. However, the number of neighboring nodes is much smaller compared with those non-neighboring nodes, which renders the model overemphasize the differences between the target and non-neighboring nodes, possibly leading to imprecise class boundaries. To solve this data-imbalance problem, we modify \emph{Mixup} \citep{zhang2017mixup}, an effective data augmentation that performs interpolation between samples to generate new training samples, to augment neighboring nodes. Specifically, we performe \textit{learnable interpolation} between the target node $v_i$ and its neighboring node $v_j\in\mathcal{N}_i$ to generate a new node, with its node representation defined as
\begin{equation}
\mathbf{z}^\prime_{i,j} = g_\gamma\big(\beta_{i,j}\mathbf{h}_j^{(L)}+(1-\beta_{i,j})\mathbf{h}_i^{(L)}\big), \quad \text{where} \ \ \  \beta_{i,j}=\mathrm{sigmoid}\big(\mathbf{a}^T \big[\mathbf{x}_i\mathbf{W}_m\|\mathbf{x}_j\mathbf{W}_m\big]\big)
\label{equ:3}
\end{equation}
where $\mathbf{W}_m \in \mathbb{R}^{d \times F}$ and $\beta_{i,j}$ is defined as \textit{learnable interpolation coefficients} with the shared attention weight $\mathbf{a}$. Then, we take augmented neighboring nodes as positive samples and other non-neighboring nodes as negative samples to simultaneously model the connectivity and disconnectivity between nodes. Specifically, the learning objective of feature-level self-distillation is defined as
\begin{equation}
\mathcal{L}_{feat}=\frac{1}{N}\sum_{i=1}^N\Big(\frac{1}{|\mathcal{N}_i|}\sum_{j\in\mathcal{N}_i} \big\|\mathbf{y}_i-\mathbf{z}^\prime_{i,j}\big\|_2^2-\frac{1}{M_i}\sum_{e_{i,k}\notin\mathcal{E}} \big\|\widehat{\mathbf{y}}_i-\widehat{\mathbf{z}}_k\big\|_2^2\Big)
\label{equ:4}
\end{equation}
where $\mathbf{z}_k=g_\gamma(\mathbf{h}_k^{(L)})$, and $M_i=|\mathcal{E}|-|\mathcal{N}_i|-1$ is the number of negative samples (non-neighborhood nodes) of the target node $v_i$. The objective $\mathcal{L}_{feat}$ essentially encourages positive neighboring nodes to be closer and pushes negative non-neighboring nodes away. Moreover, we have demonstrated the benefits of mixup-like augmentation and negative samples by the ablation study in Sec.~\ref{sec:5.4}.

\textbf{Label-level Self-Distillation.}
Thus far, we have only discussed how to exploit structural information for feature-level self-distillation, but have not explored how to leverage label information. A widely used solution to leverage label information is to optimize the objective on the labeled data $\mathcal{V}_L$ as
\begin{equation}
\min_{\theta,\mathbf{W}^{(0)},\cdots,\mathbf{W}^{(L-1)}} \sum_{i\in\mathcal{V}_L} \mathcal{L}_{CE}(s_i, \widehat{\mathbf{y}}_i), \quad \text{where} \ \ \ \widehat{\mathbf{y}}_{i}=\textrm{softmax}(\mathbf{y}_i) \in \mathbb{R}^C
\label{equ:5}
\end{equation}
where $\mathcal{L}_{CE}(\cdot)$ denotes the cross-entropy loss between $\widehat{\mathbf{y}}_{i}$ and ground-truth label $s_i$ of node $v_i$. However, Eq.~(\ref{equ:5}) only considers node labels, but completely ignores the graph structure. In practice, label propagation has been widely used as an effective trick to simultaneously model label and structural information and achieved promising results for various GNNs. However, label propagation involves explicit coupling of labels with the structure, so it is heavily data-dependent with the same inference-latency problem as message passing. Earlier, we have proposed feature-level self-distillation in Eq.~(\ref{equ:4}) to substitute message passing in Eq.~(\ref{equ:1}), and next we introduce \textit{implicit} label-level self-distillation of Eq.~(\ref{equ:6}) to substitute \textit{explicit} label propagation to fully exploit both label and structural information. The objective of label-level self-distillation can be defined as follows
\begin{equation}
\mathcal{L}_{label} = \sum_{i\in\mathcal{V}_L} \Big(\mathcal{L}_{CE}(s_i, \widehat{\mathbf{y}}_i) + \sum_{j\in\mathcal{N}_i} \mathcal{L}_{CE}(s_i, \widehat{\mathbf{z}}_j)\Big)
\label{equ:6}
\end{equation}
where $\widehat{\mathbf{y}}_{i}\!=\!\textrm{softmax}(\mathbf{y}_i) \in \mathbb{R}^C$, $\widehat{\mathbf{z}}_{j}\!=\!\textrm{softmax}(\mathbf{z}_j) \in \mathbb{R}^C$, and $s_i$ is the ground-truth label of node $v_i$.

\subsection{Training and Inferring}
\textbf{Model Training.}
After introducing two important objective functions in Eq.~(\ref{equ:4})(\ref{equ:6}), we start to consider the optimization difficulties and strategies in training. In practice, directly optimizing Eq.~(\ref{equ:4})(\ref{equ:6}) faces two tricky challenges: \textit{(1)} it treats all non-neighboring nodes as negative samples, which suffers from a huge computational burden; and \textit{(2)} it performs the summation over the entire set of nodes, i.e, requiring a large memory space for keeping the entire graph. To address these two problems, we adopt the edge sampling strategy \citep{mikolov2013distributed,tang2015line} instead of feeding the entire graph into the memory for \emph{batch-style training}. More specifically, we first sample a mini-batch edges from the entire edge set $\mathcal{E}_b\in\mathcal{E}$. Then we randomly sample negative nodes from a pre-defined negative distribution $P_k(v)$ for each edge $e_{i,j}\in\mathcal{E}_b$ instead of enumerating all non-neighboring nodes as negative samples. Finally, we can rewrite Eq.~(\ref{equ:4}) as follows
\begin{equation}
\begin{small}
\begin{aligned}
\mathcal{L}_{feat}  \! =\! \frac{1}{B}\sum_{b=1}^B  \sum_{e_{i,j}\in\mathcal{E}_b}  \bigg(
\big\|\mathbf{y}_i-\mathbf{z}^\prime_{i,j}\big\|_2^2
+\big\|\mathbf{y}_j-\mathbf{z}^\prime_{j,i}\big\|_2^2
-\mathbb{E}_{{v_k} \sim P_k(v)} \Big(\big\|\widehat{\mathbf{y}}_i-\widehat{\mathbf{z}}_k\big\|_2^2\big)
+\big\|\widehat{\mathbf{y}}_j-\widehat{\mathbf{z}}_k\big\|_2^2\big)\Big)
\bigg)
\end{aligned}
\end{small}
\label{equ:7}
\end{equation}
where $B$ is the batch size, and $P_k(v)$ adopts the uniform distribution by default, that is $P_k(v_i)=\frac{1}{N}$ for each node $v_i$. $P_k(v)$ can also be pre-defined based on prior knowledge, e.g., degree distribution, edge number, but in practice we find from the experimental results in Section.~\ref{sec:5.4} that uniform distribution is a reasonable choice that can yield fairly good performance across various datasets. Similarly, we can rewrite the label-level self-distillation of Eq.~(\ref{equ:6}) as a batch-style form, as follows
\begin{equation}
\mathcal{L}_{label}=\frac{1}{B}\sum_{b=1}^B\sum_{i\in\mathcal{V}_L\cap\mathcal{V}_b}
\Big(\mathcal{L}_{CE}(s_i, \widehat{\mathbf{y}}_i) + \sum_{e_{i,j}\in\mathcal{E}_b} \mathcal{L}_{CE}(s_i, \widehat{\mathbf{z}}_j)\Big)
\label{equ:8}
\end{equation}
where $\mathcal{V}_b=\{v_i,v_j|e_{i,j}\in\mathcal{E}_b\}$ is all the sampled nodes in $\mathcal{E}_b$. Finally, the total loss for training the model can be defined as $\mathcal{L}_{total}=\mathcal{L}_{label}+\lambda\mathcal{L}_{feat}$, where $\lambda$ is a trade-off hyperparameter.

\textbf{Model Inferring.}
Once the model training is completed, we can directly omit the inference layer $g_\gamma(\cdot)$ and retain the backbone MLP architecture and the inference layer $f_\theta(\cdot)$ for label prediction. At this time, there is no data dependency for model inference, and it is attributed to the fact that we have shifted a considerable amount of work from the latency-sensitive inference stage to the latency-insensitive training stage. The pseudo-code is summarized in Algorithm~\ref{algo:1} in \textbf{Appendix A}.

\subsection{Discussion and Comparison}
In this subsection, we would like to compare the proposed GSDN framework with some existing related work in order to make a better distinction and highlight our contributions.

\textbf{Comparison with message and label propagation.}
The core of GNNs is the use of structural information, where \emph{message passing} and \emph{label propagation} are the two dominant schemes. The message passing models the long-range data dependency between nodes through neighborhood feature aggregation, while label propagation focuses on diffusing label information to the neighborhood, and \emph{they are complementary to each other}. However, both of them involve explicit coupling of features/labels with structures, leading to data dependency and inference latency. Different from the \textit{explicit} message passing and label propagation, we use structural information as prior, as shown in Fig.~\ref{fig:2}, to \textit{implicitly} guide neighborhood feature-level self-distillation (from neighborhood to the target node) as in Eq.~(\ref{equ:4}), and label-level self-distillation (from the target node to the neighborhood) as in Eq.~(\ref{equ:6}), where structural information is never explicitly involved in the forward propagation. Furthermore, we have rethinked the differences between GNNs and GSDN from the perspective of information flow, explaining how GSDN defined on the 1-hop neighborhood influence nodes that are multi-hops away; however, due to space limitations, we place this part of the discussion in \textbf{Appendix B}.

\begin{wrapfigure}{r}{0.53\textwidth}
\vspace{-1em}
\begin{center}
\includegraphics[width=0.55\textwidth]{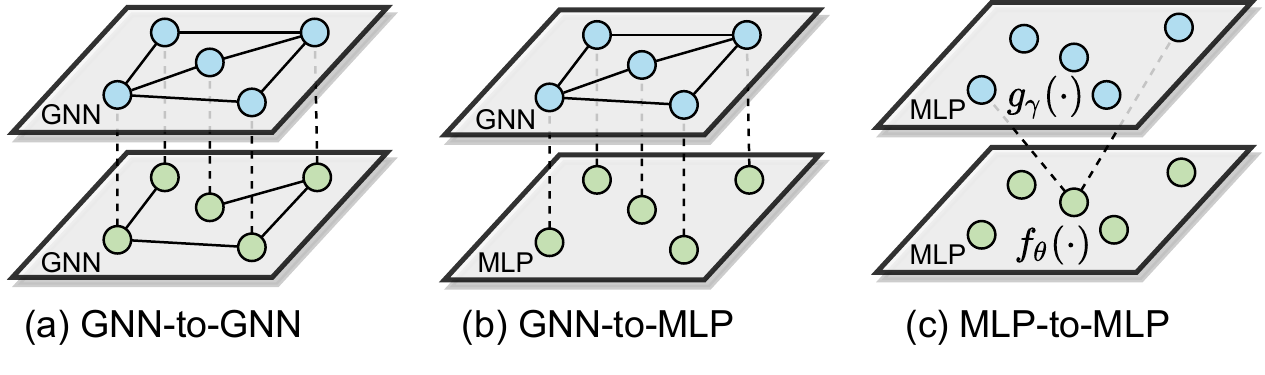}   
\end{center}
\vspace{-1.5em}
\caption{A comparison of three distillation methods.}
\label{fig:3}
\end{wrapfigure}

\textbf{Discussion on knowledge distillation.}
Several previous works on GNN distillation seek to distill knowledge from large teacher GNN to more lightweight student models. Potential lightweight student models that could be employed could be a GNN with fewer parameters, termed GNN-to-GNN distillation \citep{yang2020distilling,yan2020tinygnn}; or it could be a pure MLP, called GNN-to-MLP distillation \citep{yang2021extract,zhang2021graph}. In contrast to the teacher-student knowledge distillation schemes, this paper adopts knowledge self-distillation, which regularizes a model by distilling its own knowledge without other teacher models. In the proposed framework, we extract representations with a shared MLP backbone, then use $g_\gamma(\cdot)$ and $f_\theta(\cdot)$ for label prediction on neighboring nodes and target node, and finally train the model by regularizing their label distribution consistencies, which can be considered as a special version of MLP-to-MLP distillation. A comparison of these three graph knowledge distillation methods is shown in Fig.~\ref{fig:3}.

\vspace{-1em}
\section{Experiments}
\vspace{-1em}
In this section, we evaluate GSDN on six real-world datasets by answering the following five questions. \textbf{Q1}: How does GSDN compare with GNN- and MLP-based models? \textbf{Q2}: Is GSDN robust under limited labeled data and label noise? \textbf{Q3}: How do GSDN compare with other general GNN models and inference acceleration methods in terms of inference time? \textbf{Q4}: How does GSDN benefit from negative samples, mixup-like augmentation, and feature/label self-distillation? \textbf{Q5}: How do the two key hyperparameters, trade-off weight $\lambda$ and batch size $B$, influence the model performance?

\textbf{Datasets.} The experiments are conducted on six widely used real-world datasets, including Cora \citep{sen2008collective}, Citeseer \citep{giles1998citeseer}, Coauthor-CS, Coauthor-Physics, Amazon-Com, and Amazon-Photo \citep{shchur2018pitfalls}. For each dataset, following the data splitting settings of \citep{kipf2016semi}, we select 20 nodes per class to construct a training set, 500 nodes for validation, and 1000 nodes for testing. A statistical overview of these six datasets is placed in \textbf{Appendix C}.

\textbf{Baseline.} We consider the following six classical baselines: Vanilla MLP, GCN \citep{kipf2016semi}, GAT \citep{velivckovic2017graph}, GraphSAGE \citep{hamilton2017inductive}, APPNP \citep{klicpera2018predict}, and DAGNN \citep{liu2020towards}. In addition, we compare the proposed framework with \textbf{two pure MLP-based architectures}, including Graph-MLP \citep{hu2021graph} and LinkDist. Besides, \textbf{five graph distillation methods} are considered as baselines: RDD \citep{zhang2020reliable}, TinyGNN \citep{yan2020tinygnn}, CPR \citep{yang2021extract}, GLNN \citep{zhang2021graph}, and GNN-SD \citep{chen2020self}. Moreover, with GCN as the base architecture, \textbf{three inference acceleration methods} are compared, including pruning with 50\% weights (P-GCN) \citep{han2015learning}, quantization from FP32 to INT8 (Q-GCN) \citep{gupta2015deep}, neighbor sampling with fan-out 15 (NS-GCN) \citep{chen2018fastgcn}. Note that there is so much work related to this paper that it would be impractical to compare with all of them, so we have selected only those classical works from various fields for comparison.

\textbf{Hyperparameter.} The hyperparameters are set the same for all datasets: Adam optimizer with learning rate $\alpha$ = 0.01 and weight decay $decay$ = 5e-4; Epoch $E$ = 200; Layer number $L$ = 2. The other dataset-specific hyperparameters are determined by an AutoML toolkit NNI with the hyperparameter search spaces as: hidden dimension $F=\{256, 512, 1024\}$; batch size $B=\{256, 512, 1024, 4096\}$, trade-off weight $\lambda=\{0.5, 0.8, 1.0\}$. Each set of experiments is run five times with different random seeds, and the average accuracy and standard deviation are reported as metrics. Moreover, the experiments of both baselines and our approach are implemented based on the standard implementation in the DGL library \citep{wang2019dgl} using the PyTorch 1.6.0 library with Intel(R) Xeon(R) Gold 6240R @ 2.40GHz CPU and NVIDIA V100 GPU. The hyperparameter sensitivity analysis (\textbf{Q5}) for the trade-off weight $\lambda$ and batch size $B$ is available in \textbf{Appendix D}.

\vspace{-0.5em}
\subsection{Performance Comparison (Q1)}
\vspace{-0.5em}
To answer \textbf{Q1}, we conduct experiments on six real-world datasets with comparison to state-of-the-art methods, three types of which are included: general GNNs, GNN distillation, and MLP-based models. Table.~\ref{tab:1} reports the mean classification accuracy with the standard deviation on the test nodes, from which we can observe that: \textit{(1)} While existing MLP-based models, such as Graph-MLP and LinkDist, can achieve comparable performance to GCN on a few datasets, they still lag far behind state-of-the-art GNNs, such as APPNP and DAGNN, and cannot even match the performance of GraphSAGE and GAT on some datasets. \textit{(2)} Regarding the graph distillation methods, while the performance of the distilled student model is significantly improved over the teacher model on some datasets, such improvement becomes less significant on a few large-scale datasets, such as Coauthor-Phy and Amazon-Photo. \textit{(3)} In terms of classification accuracy, GSDN consistently achieves the best overall performance on six datasets, even better than the state-of-the-art GNN models, such as APPNP and DAGNN. For example, GSDN obtains the best performance on the Coauthor-Physics dataset, and more notably, our accuracy outperforms DAGNN by 1.86\%, which once again demonstrates the effectiveness of the GSDN framework for the node classification task.

\begin{table*}[!tbp]
\begin{center}
\vspace{-2em}
\caption{Classification accuracy $\pm$ std (\%), with the best and second results marked by \textbf{bold} and \underline{underline}. If not specifically mentioned, all relevant models adopt GCN as the backbone by default.}
\vspace{-0.5em}
\label{tab:1}
\resizebox{1.0\textwidth}{!}{
\begin{tabular}{clccccccc}

\toprule
\textbf{Type} & \textbf{Method} & \textbf{Cora} & \textbf{Citeseer} & \textbf{Coauthor-CS} & \textbf{Coauthor-Phy} & \textbf{Amazon-Com} & \textbf{Amazon-Photo} \\ \midrule
\multirow{5}{*}{General GNNs} & GCN & 81.28$\pm$0.42 & 71.06$\pm$0.44 & 87.76$\pm$0.43 & 91.89$\pm$0.42 & 77.45$\pm$1.71 & 87.53$\pm$1.64 \\
 & GAT & 83.02$\pm$0.45 & 72.56$\pm$0.51 & 88.55$\pm$0.56 & 92.36$\pm$0.47 & \underline{82.78$\pm$1.89} & 90.19$\pm$1.35 \\
 & GraphSAGE & 82.22$\pm$0.80 & 71.22$\pm$0.58 & 88.40$\pm$0.48 & 91.88$\pm$0.53 & 79.23$\pm$1.63 & 88.63$\pm$1.17 \\
 & APPNP & 83.28$\pm$0.33 & 71.74$\pm$0.27 & 88.74$\pm$0.62 & 92.75$\pm$0.60 & 81.28$\pm$1.90 & 89.49$\pm$1.28 \\
 & DAGNN & 84.30$\pm$0.51 & 73.14$\pm$0.62 & 89.32$\pm$0.55 & \underline{93.10$\pm$0.67} & 80.32$\pm$1.57 & \underline{90.72$\pm$1.45} \\ \midrule
 
\multirow{4}{*}{Graph Distillation} & RDD & \underline{84.68$\pm$0.40} & \underline{73.63$\pm$0.50} & \underline{89.38$\pm$0.44} & 92.74$\pm$0.78 & 81.84$\pm$1.48 & 89.70$\pm$0.93 \\
 & TinyGNN & 82.79$\pm$0.57 & 72.67$\pm$0.72 & 88.72$\pm$0.42 & 92.20$\pm$0.67 & 79.22$\pm$1.69 & 89.24$\pm$1.24 \\ 
 & CPR & 83.65$\pm$0.49 & 72.98$\pm$0.47 & 89.10$\pm$0.50 & 92.36$\pm$0.63 & 80.90$\pm$1.52 & 89.03$\pm$1.29 \\
 & GLNN & 80.85$\pm$0.60 & 71.21$\pm$0.80 & 87.81$\pm$0.53 & 91.83$\pm$0.60 & 77.96$\pm$1.70 & 87.98$\pm$1.36 \\ 
 & GNN-SD & 81.85$\pm$0.55 & 71.69$\pm$0.61 & 87.80$\pm$0.50 & 92.07$\pm$0.48 & 77.66$\pm$1.85 & 87.80$\pm$1.52 \\ \midrule
 
\multirow{4}{*}{MLP-based} & MLP & 61.86$\pm$0.43 & 59.76$\pm$0.51 & 83.34$\pm$0.64 & 86.24$\pm$0.66 & 66.85$\pm$1.94 & 78.18$\pm$1.25 \\
 & Graph-MLP & 81.45$\pm$0.52 & 72.87$\pm$0.70 & 88.16$\pm$0.70 & 91.85$\pm$0.49 & 77.23$\pm$1.76 & 87.64$\pm$1.37 \\
 & LinkDist & 76.70$\pm$0.47 & 65.19$\pm$0.55 & 87.89$\pm$0.58 & 92.16$\pm$0.70 & 76.93$\pm$1.83 & 87.26$\pm$1.42 \\
 & GSDN (ours) & \textbf{84.90$\pm$0.44} & \textbf{74.08$\pm$0.69} & \textbf{89.62$\pm$0.40} & \textbf{94.96$\pm$0.41} & \textbf{83.44$\pm$2.09} & \textbf{90.34$\pm$0.85} \\ \bottomrule
 
\end{tabular}} \vspace{-1.5em}
\end{center}
\end{table*}

\vspace{-0.5em}
\subsection{Evaluation on Robustness (Q2)}
\vspace{-0.5em}
There has been some work pointing out that the performance of GNNs depends heavily on the quality and quantity of the labels. To evaluate the robustness of the proposed framework, we evaluate the model with extremely limited label data and under label noise on the Cora and Citeseer datasets.

\begin{table}[!htbp]
\begin{center}
\vspace{-1em}
\caption{Accuracy $\pm$ std (\%) with extremely limited labels, with the best results marked by \textbf{bold}.}
\vspace{-0.5em}
\label{tab:2}
\resizebox{\textwidth}{!}{
\begin{tabular}{cccccc|cccc|cc}

\toprule
\multicolumn{2}{c}{Dataset}           & GCN        & GAT        & APPNP      & DAGNN      & RDD        & CPR        & GLNN       & GNN-SD     & Graph-MLP  & GSDN (ours)         \\ \midrule
\multirow{3}{*}{Cora}     & 5 labels  & 73.10$\pm$0.87 & 74.85$\pm$0.74 & 76.39$\pm$0.95 & 79.02$\pm$0.94 & 76.11$\pm$0.81 & 75.92$\pm$0.90 & 73.85$\pm$0.92 & 74.32$\pm$0.85 & 78.43$\pm$0.78 & \textbf{80.22$\pm$0.87} \\
                          & 10 labels & 77.52$\pm$0.63 & 80.10$\pm$0.57 & 79.99$\pm$0.72 & 81.99$\pm$0.62 & 79.68$\pm$0.54 & 79.20$\pm$0.64 & 77.94$\pm$0.52 & 78.55$\pm$0.64 & 79.60$\pm$0.49 & \textbf{82.80$\pm$0.45} \\
                          & 15 labels & 79.47$\pm$0.56 & 80.55$\pm$0.68 & 80.70$\pm$0.44 & 82.72$\pm$0.50 & 80.45$\pm$0.46 & 80.12$\pm$0.51 & 79.16$\pm$0.48 & 79.89$\pm$0.55 & 80.32$\pm$0.57 & \textbf{83.40$\pm$0.53} \\ \midrule
\multirow{3}{*}{Citeseer} & 5 labels  & 63.23$\pm$1.04 & 64.17$\pm$0.95 & 66.28$\pm$0.88 & 69.13$\pm$0.68 & 66.06$\pm$0.57 & 65.40$\pm$0.62 & 63.10$\pm$0.70 & 63.93$\pm$0.74 & \textbf{69.64$\pm$0.64} & 68.76$\pm$1.16 \\
                          & 10 labels & 67.55$\pm$0.50 & 68.09$\pm$0.49 & 69.23$\pm$0.64 & 71.74$\pm$0.71 & 69.68$\pm$0.66 & 69.14$\pm$0.73 & 67.65$\pm$0.62 & 68.20$\pm$0.57 & 70.56$\pm$0.51 & \textbf{72.66$\pm$0.43} \\
                          & 15 labels & 69.64$\pm$0.58 & 69.70$\pm$0.65 & 70.17$\pm$0.44 & 72.26$\pm$0.53 & 70.32$\pm$0.57 & 70.76$\pm$0.46 & 69.52$\pm$0.52 & 69.86$\pm$0.48 & 71.80$\pm$0.63 & \textbf{73.10$\pm$0.53} \\ \bottomrule
                          
\end{tabular}} \vspace{-0.5em}
\end{center}
\end{table}

\textbf{Performance with Severely Limited Labels.} To evaluate the effectiveness of the proposed GSDN framework when labeled data is severely limited, we randomly select 5, 10, and 15 labeled samples per class for training, and the rest of the training set is considered unlabeled. The classification performance on the Cora and Citeseer datasets is reported in Table.~\ref{tab:2}, from which we can observe that (1) The performance of all methods drops as the number of labeled data decreases, but the accuracy of GSDN drops more slightly. When only a limited number of labels are provided, GSDN outperforms all other baselines at most label rates. For example, when trained with five labels per class, GSDN outperforms GCN by 7.12\% and 5.53\% on the Cora and Citeseer datasets. (2) While graph distillation methods perform well on clean data, as shown in Table.~\ref{tab:1}, their performance gains are reduced when labeled data is extremely limited. In contrast, MLP-based models, both Graph-MLP and GSDN, show great advantages over GNN-based models under the label-limited setting.

\begin{figure}[!tbp]
    \vspace{-2em}
	\begin{center}
		\subfigure[Performance on Cora]{\includegraphics[width=0.315\linewidth]{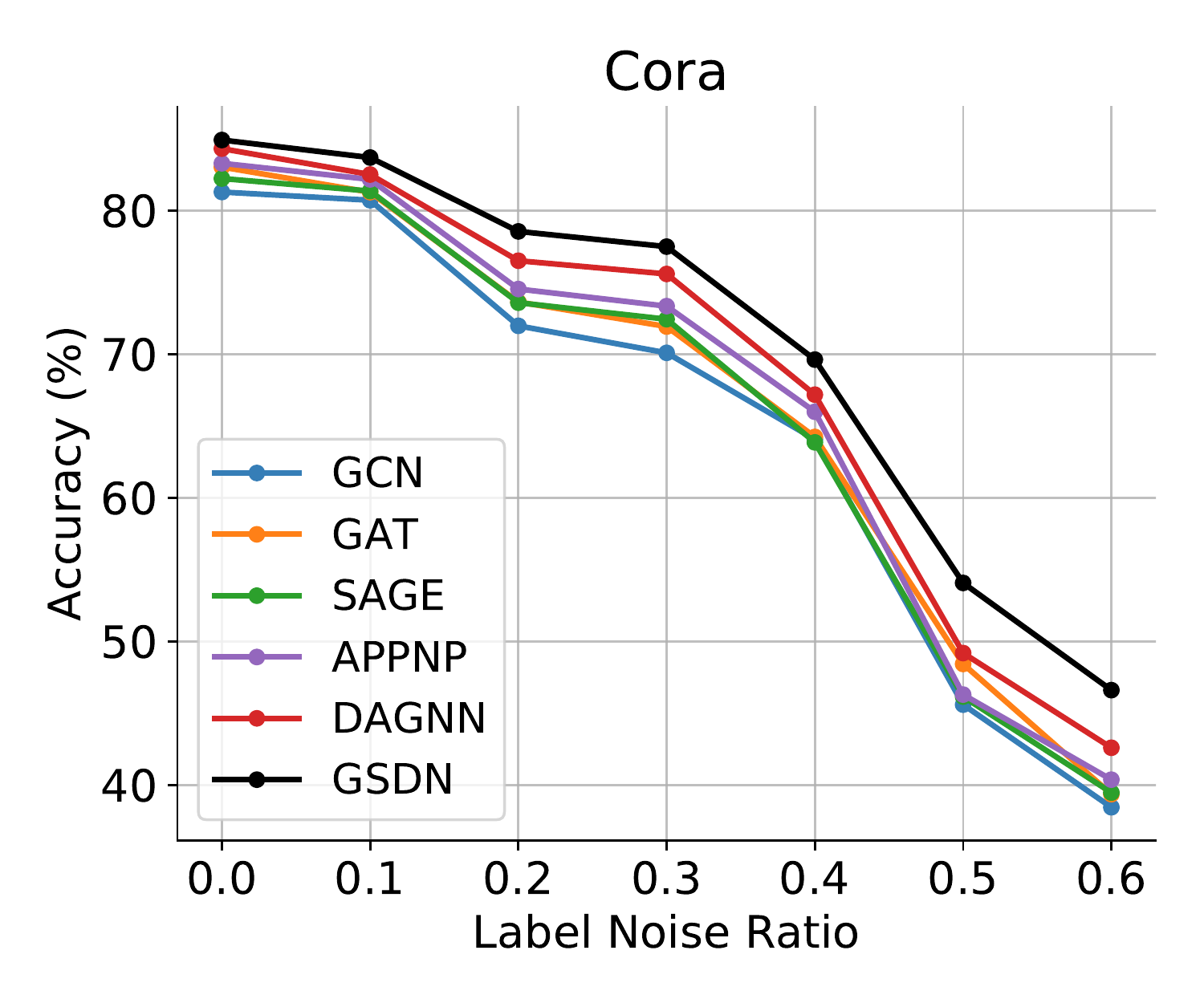}\label{fig:4a}}
		\subfigure[Performance on Citeseer]{\includegraphics[width=0.315\linewidth]{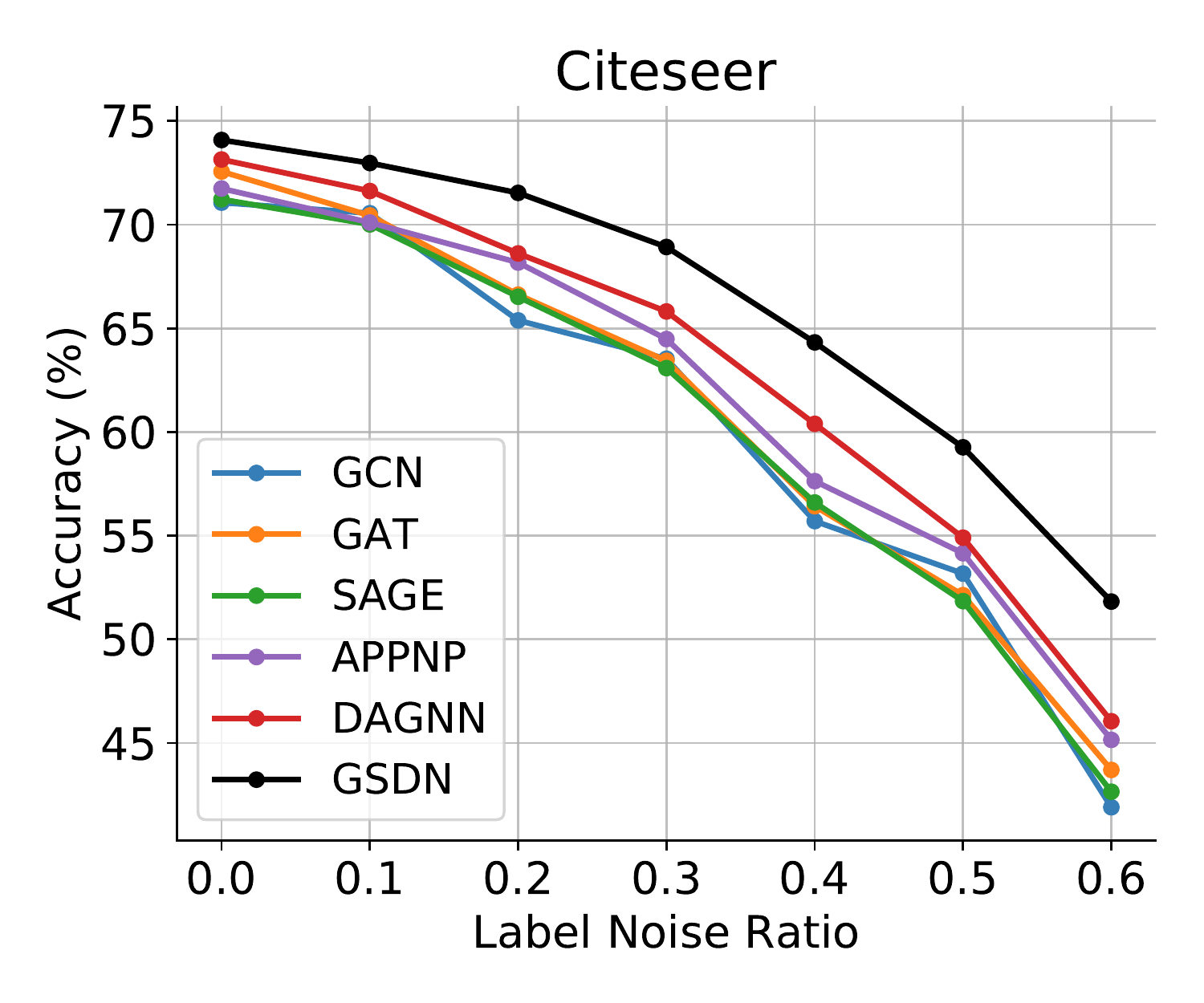}\label{fig:4b}}
		\subfigure[Inference time with different layers]{\includegraphics[width=0.35\linewidth]{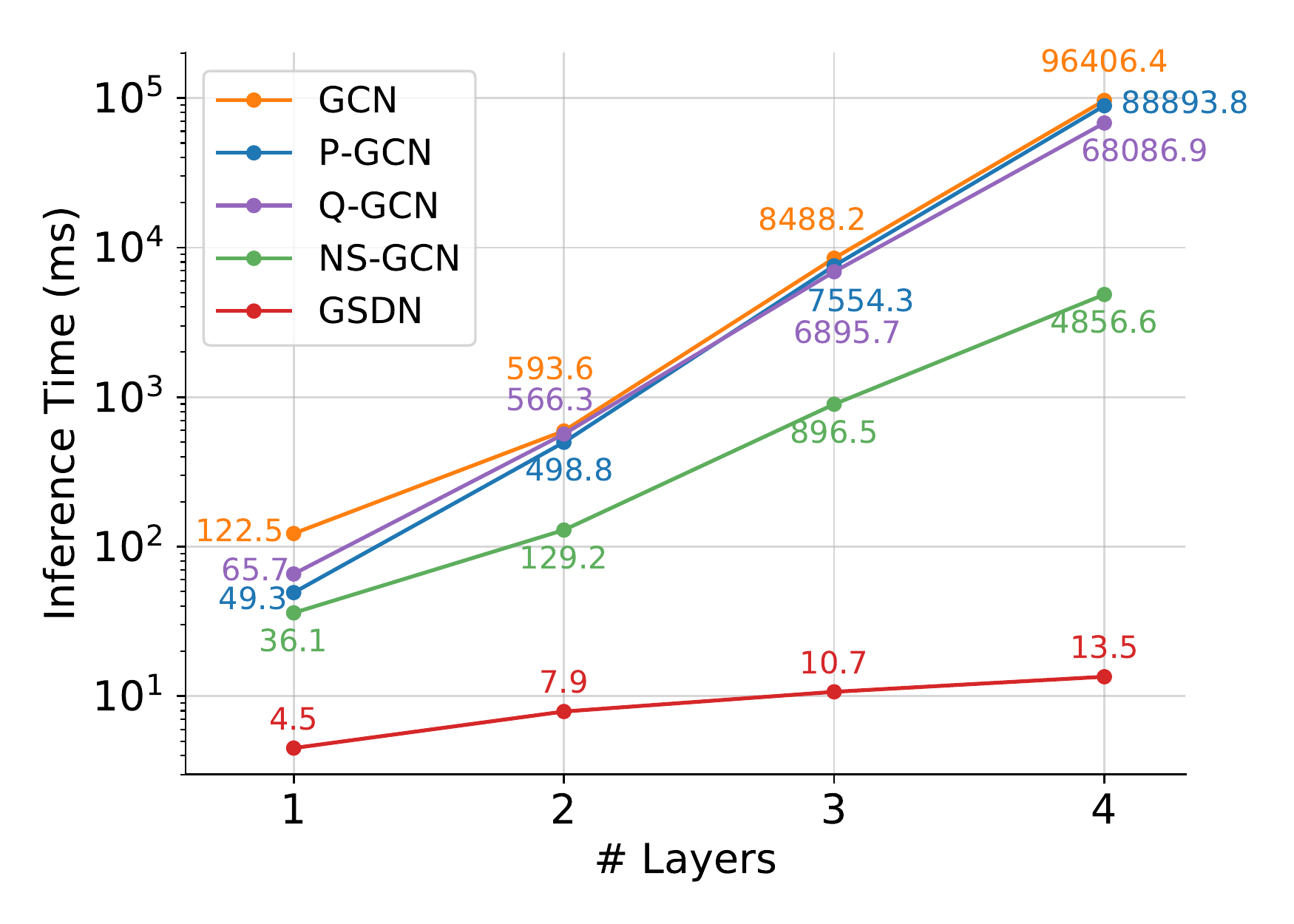}\label{fig:4c}}
	\end{center}
	\vspace{-1em}
	\caption{(a)(b) Classification accuracy (\%) under different label noise ratios on the Cora and Citeseer datasets, respectively. (c) Inference time with different layers on the Coauthor-CS dataset.}
	\vspace{-1.5em}
	\label{fig:4}
\end{figure}

\textbf{Performance with Noisy Labels.} We evaluate the robustness of GSDN to label noise by injecting \textit{asymmetric} noise into the labels, where the label $i$ $(0\leq i \leq C-1)$ of each training sample flips independently with probability $r$ to another class $j = (i+1) \% C $, but with probability $1-r$ preserved as label $i$ \citep{tan2021co}. The performance with noisy labels is reported in Fig.~\ref{fig:4a} and Fig.~\ref{fig:4b} at various noise ratios $r\in\{0\%, 10\%, 20\%, \cdots 60\%\}$. It can be seen that as $r$ increases, the accuracy of GSDN drops more slowly than other baselines, and GSDN is more robust than other models under various noise ratios, especially under extremely high noise ratios. For example, with $r=60\%$ label noise, GSDN outperforms APPNP and DAGNN by 6.67\% and 5.77\% on the Citeseer dataset, respectively. Due to space limitations, more evaluation results on noisy labels, including comparisons with other graph distillation and MLP-based models, are available in \textbf{Appendix E}.

\vspace{-0.5em}
\subsection{Evaluation on Inference Speed (Q3)} \label{sec:5.3}
\vspace{-0.5em}
Commonly used inference acceleration techniques on GNNs include pruning, quantizing, and neighbor sampling. With GCN as the backbone, we consider its three variants: P-GCN, Q-GCN, and NS-GCN. A comparison with more approaches, such as graph distillation and MLP-based models, in terms of inference speed, can be seen in Fig.~\ref{fig:1c} and \textbf{Appendix F}. Moreover, since the focus of this paper is on inference speed, we defer the results and analysis on training time to \textbf{Appendix G}.

\textbf{Comparison on Different Datasets.}
With the removal of neighborhood fetching, the inference time of GSDN can be reduced from $\mathcal{O}(|\mathcal{V}|dF+|\mathcal{E}|F)$ of GCN to $\mathcal{O}(|\mathcal{V}|dF)$. The inference time ($ms$) averaged over 30 sets of runs on four datasets is reported in Table.~\ref{tab:3} with the acceleration multiple w.r.t the vanilla GCN marked as ${\color[rgb]{0.4, 0.71, 0.376}green}$, where all methods use $L=2$ layers and hidden dimension $F=16$.  From Table.~\ref{tab:3}, we have the following observations: \textit{(1)} While APPNP and DAGNN improve a lot over GCN in terms of classification accuracy, as shown in Table.~\ref{tab:1}, they suffer from more severe inference latency. (2) Pruning and quantization are not very effective on GNNs, given that data dependency has not been well resolved. Besides, the neighbor sampling considers but does not completely eliminate the neighborhood-fetching latency, so it infers faster than pruning and quantization, yet still lags far behind our GSDN. (3) GSDN infers fastest across four datasets.

\begin{table}[!htbp]
\begin{center}
\vspace{-1em}
\caption{Inference time ($ms$), where three inference acceleration methods speed up GCN, but still infer slower than GSDN. Note the acceleration multiple w.r.t the vanilla GCN is marked as ${\color[rgb]{0.4, 0.71, 0.376}green}$.}
\vspace{-0.5em}
\label{tab:3}
\resizebox{\columnwidth}{!}{
\begin{tabular}{lccccccc}

\toprule
\textbf{Method} & GCN & APPNP & DAGNN & P-GCN & Q-GCN & NS-GCN & GSDN (ours) \\ \midrule
\textbf{Cora} & 402.7 & 731.8 & 647.6 & 372.9 (${\color[rgb]{0.4, 0.71, 0.376}1.08\times}$) & 383.5 (${\color[rgb]{0.4, 0.71, 0.376}1.05\times}$) & 97.7 (${\color[rgb]{0.4, 0.71, 0.376}4.12\times}$) & 4.6 (${\color[rgb]{0.4, 0.71, 0.376}87.54\times}$) \\
\textbf{Citeseer} & 383.4 & 700.5 & 727.5 & 316.9 (${\color[rgb]{0.4, 0.71, 0.376}1.21\times}$) & 361.4 (${\color[rgb]{0.4, 0.71, 0.376}1.06\times}$) & 105.6 (${\color[rgb]{0.4, 0.71, 0.376}3.63\times}$) & 4.3 (${\color[rgb]{0.4, 0.71, 0.376}89.16\times}$) \\
\textbf{Coauthor-CS} & 593.6 & 1099.1 & 672.4 & 498.8 (${\color[rgb]{0.4, 0.71, 0.376}1.19\times}$) & 566.3 (${\color[rgb]{0.4, 0.71, 0.376}1.05\times}$) & 129.2 (${\color[rgb]{0.4, 0.71, 0.376}4.59\times}$) & 7.9 (${\color[rgb]{0.4, 0.71, 0.376}75.14\times}$) \\
\textbf{Coauthor-Phy} & 1067.6 & 1467.4 & 708.6 & 922.3 (${\color[rgb]{0.4, 0.71, 0.376}1.16\times}$) & 997.8 (${\color[rgb]{0.4, 0.71, 0.376}1.07\times}$) & 269.3 (${\color[rgb]{0.4, 0.71, 0.376}3.96\times}$) & 12.8 (${\color[rgb]{0.4, 0.71, 0.376}83.41\times}$) \\ \bottomrule

\end{tabular}} \vspace{-1em}
\end{center}
\end{table}

\textbf{Comparison with Different Layers.}
As discussed earlier, to infer a single node with a $L$-layer GNN on a graph with average node degree $R$, it requires fetching and aggregating $\mathcal{O}(R^L)$ nodes. To compare the sensitivity of GSDN to layer depth with other inference acceleration methods, we report the inference time ($ms$ in log-scale) of various methods at different layer depths on the Coauthor-CS dataset in Fig.~\ref{fig:4c}. It can be seen that the inference time of GSDN only increases \textbf{linearly} with the layer depth, but that of other baselines increases \textbf{exponentially}. Moreover, the speed gains of pruning and quantization on GNNs are reduced as the layer depth increases, and they approach the vanilla GCN when the layer depth is 4. In contrast, at larger layer depths, the speed gain of the neighbor sampling gets enlarged compared to the vanilla GCN. \textbf{This demonstrates that neighborhood fetching is one major source of inference latency} as the layer depth increases, and the linear complexity of GSDN has a significant advantage, especially when GNNs is becoming deeper.

\begin{figure*}[!tbp]
    \vspace{-2em}
	\begin{center}
        \subfigure[Ablation Study]{\includegraphics[width=0.33\linewidth]{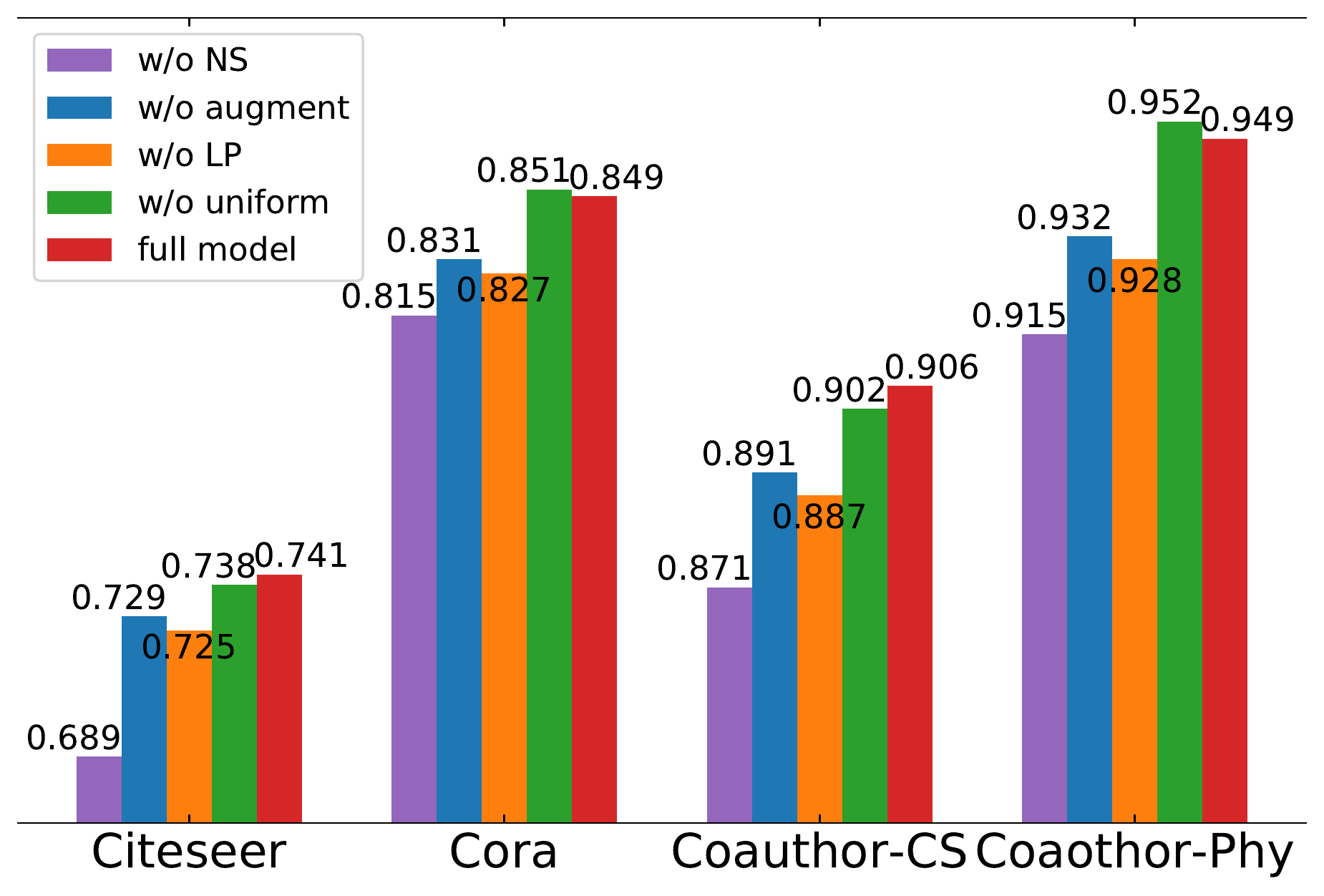}\label{fig:5a}}
		\subfigure[Learning curves]{\includegraphics[width=0.32\linewidth]{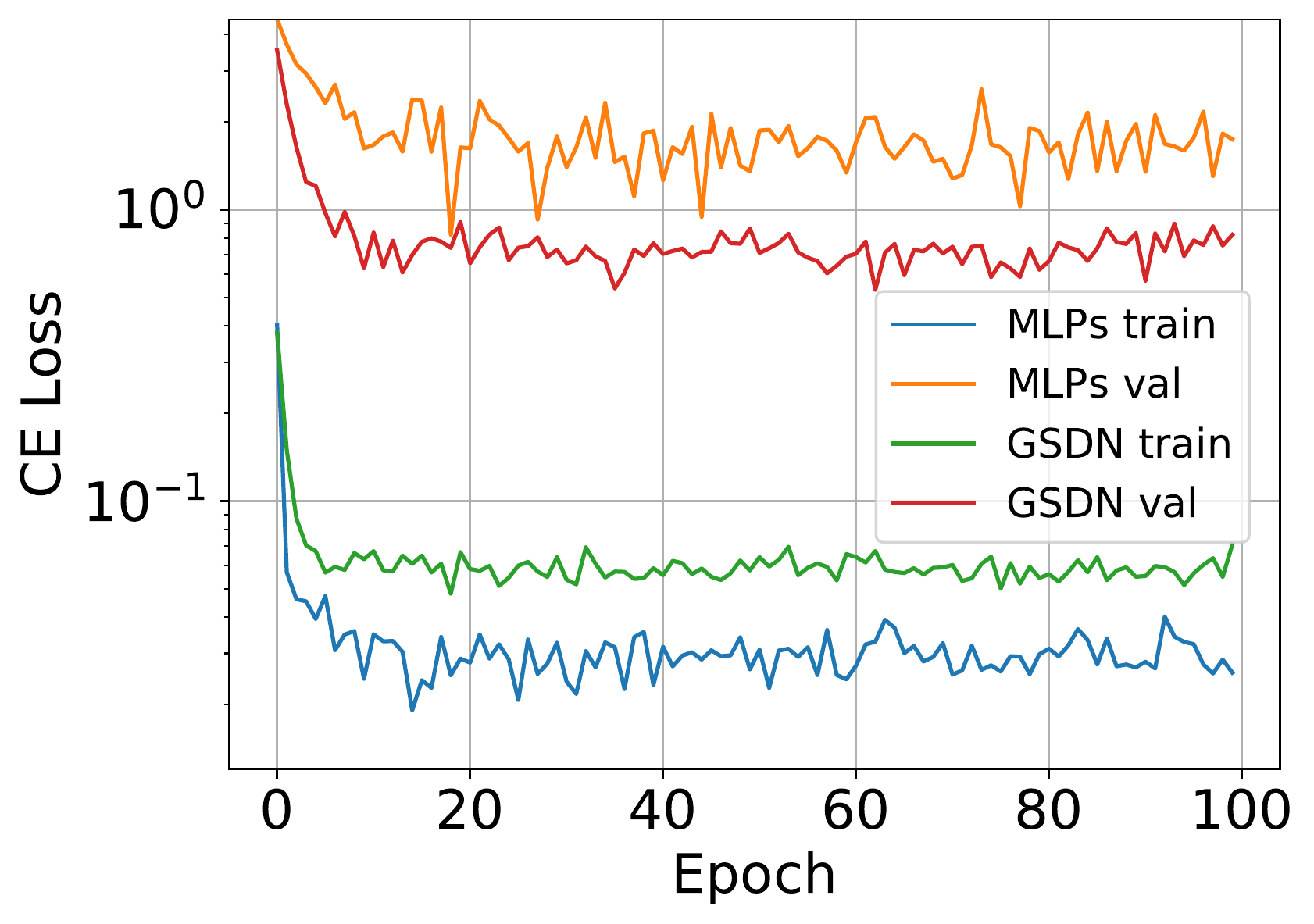}\label{fig:5b}}
		\subfigure[Mean cosine similarity curves]{\includegraphics[width=0.32\linewidth]{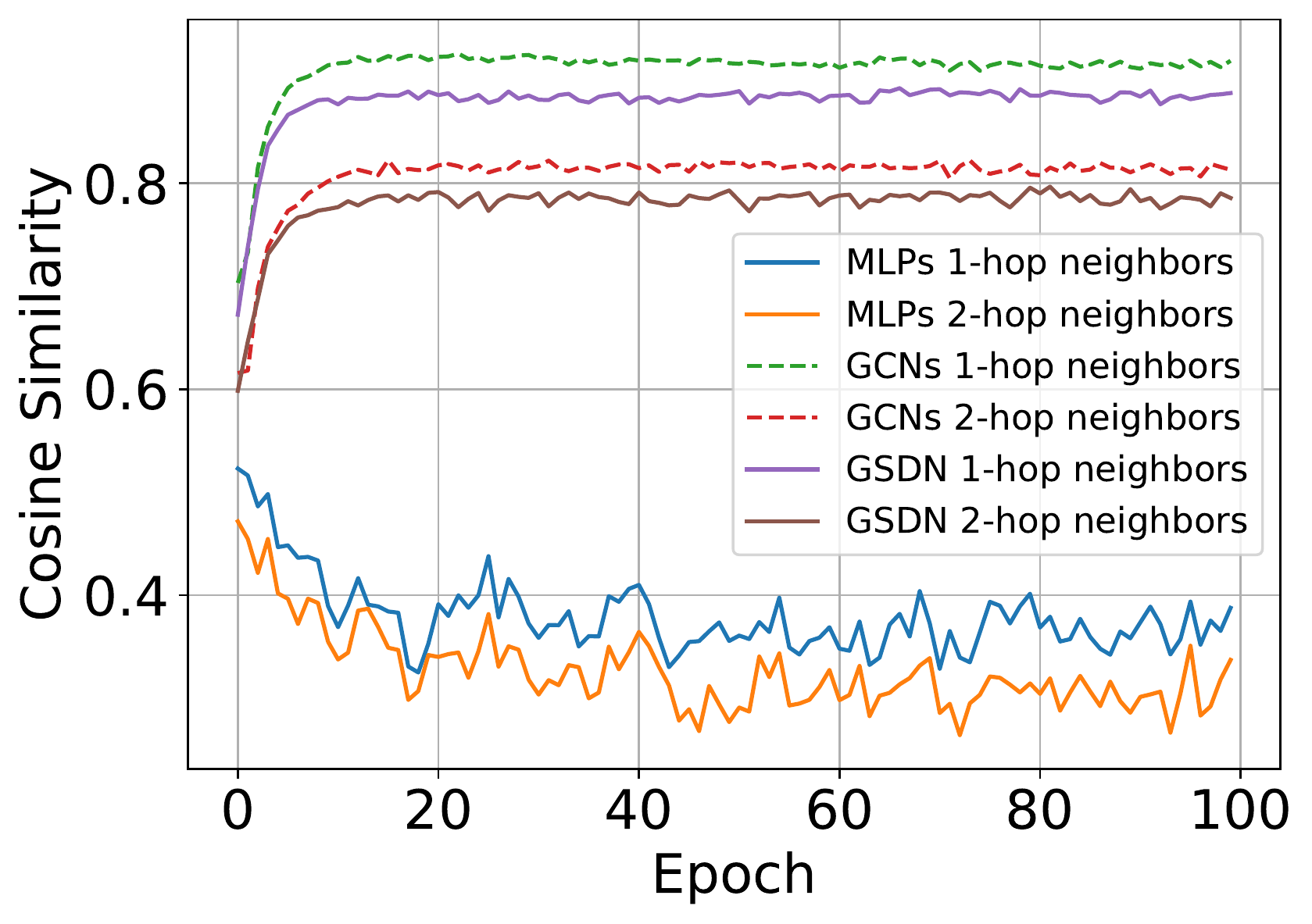}\label{fig:5c}}
	\end{center}
	\vspace{-1em}
	\caption{(a) Ablation study on four key model components. (b) Learning curves of MLPs and GSDN on the Cora dataset, where the \emph{logarithmized} vertical coordinate is the cross-entropy loss between the predicted and ground-truth labels on the training or validation set. (c) Mean cosine similarity curves of MLPs, GCNs, and GSDN with 1-hop and 2-hop neighbors on the Cora dataset.}
	\vspace{-1em}
	\label{fig:5}
\end{figure*}

\vspace{-0.5em}
\subsection{Ablation Study (Q4)} \label{sec:5.4}
\vspace{-0.5em}
\textbf{Component Analysis.}
To evaluate the effectiveness of negative samples in Eq.~(\ref{equ:4}), mixup-like augmentation in Eq.~(\ref{equ:3}), and label self-distillation in Eq.~(\ref{equ:8}), we conducted four sets of experiments: the model without (A) Negative Samples (\textit{w/o} NS); (B) mixup-like augmentation (\textit{w/o} augment); (C) Label self-Distillation (\textit{w/o} LD); and (D) the full model. Besides, to evaluate the impact of the negative distribution, we take the the nodal degree $d_i$ as a prior and preset $P_k(v_i)=\frac{d_i}{|\mathcal{E}|}$ in place of the default uniform distribution in this paper, denoted as (E) \textit{w/o} uniform. After analyzing the results in Fig.~\ref{fig:5a}, we can conclude: (1) Negative Samples and mixup-like augmentation contribute to improving classification performance. More importantly, applying them together can further improve performance on top of each. (2) Label self-distillation helps to improve performance on top of the stand-alone feature self-distillation. (3) Even without considering any graph prior, presetting $P_k(\cdot)$ as uniform distribution is sufficient to achieve comparable performance, so this paper defaults to the simplest uniform distribution without considering other complex prior-based distributions.

\textbf{How GSDN Benefit from Neighborhood Self-Distillation.}
Next, we explore how does GSDN benefit from neighborhood self-distillation? Existing MLP-based models have shown that there do exist the optimal MLP parameters that enable their performance to be competitive with GNNs on the attribute graph, but it is hard to learn such parameters through a simple cross-entropy loss \citep{hu2021graph}. The proposed self-distillation helps to solve this problem with two potential advantages: (1) alleviating overfitting and (2) introducing inductive bias, i.e., graph topology \citep{zhang2021graph}.

Firstly, we plot the training curves (with log-scale vertical coordinate) of GSDN and MLPs on the Cora dataset in Fig.~\ref{fig:5b}. We observe that the gap between training and validation loss is smaller for GSDN than MLPs, which indicates that GSDN helps to alleviate the overfitting trend of MLPs. Secondly, we conjecture that the absence of inductive bias, e.g., graph topology, is one of the major reasons why MLPs is inferior to GNNs in inference accuracy. To illustrate it, we plot in Fig.~\ref{fig:5c} the average cosine similarity of nodes with their 1-hop and 2-hop neighbors for MLPs, GCNs, and GSDN on the Cora dataset. It can be seen that the average similarity with 1-hop neighbors is always higher than that with 2-hop neighbors throughout the training process for MLPs, GCNs, and GSDN. More importantly, the average similarity of GCNs and GSDN gradually increases with training, while that of MLPs gradually decreases, which indicates that GSDN has introduced graph topology as an inductive bias (as GCNs has done), while MLPs does not. Finally, the GSDN enjoys the benefits of topology-awareness in training but without neighborhood-fetching latency in inference.

\vspace{-0.5em}
\section{Conclusion}
\vspace{-0.5em}
Motivated by the complementary strengths and weaknesses of GNNs and MLPs, we propose a novel MLP-based framework, namely \textit{Graph Self-Distillation on Neighborhood} (GSDN), where structural information is only implicitly used as prior to guide knowledge self-distillation between the neighborhood and the target. As a result, GSDN shifts a considerable amount of work from the latency-sensitive inference stage to the latency-insensitive training state, thus enjoying the benefits of graph topology-awareness in training but without data dependency in inference. More importantly, we study GSDN properties comprehensively by investigating how they benefit from neighborhood self-distillation and how they are different from existing works. Extensive experiments show the advantages of GSDN over existing methods in terms of inference accuracy and inference efficiency.

\clearpage
\bibliography{iclr2023_conference}
\bibliographystyle{iclr2023_conference}

\clearpage
\renewcommand\thefigure{A\arabic{figure}}
\renewcommand\thetable{A\arabic{table}}
\renewcommand\theequation{A.\arabic{equation}}
\setcounter{table}{0}
\setcounter{figure}{0}
\setcounter{theorem}{0}
\setcounter{equation}{0}

\section*{Appendix}

\subsection*{A. Pseudo Code of GSDN}
The pseudo-code of the proposed GSDN framework is summarized in Algorithm~\ref{algo:1}.

\begin{algorithm}[!htbp]
	\caption{Algorithm for the proposed \textit{GSDN} framework}
	\label{algo:1}
	\begin{algorithmic}[1]
		\Require Feature Matrix: $\mathbf{X}$; Edge Set: $\mathcal{E}$; \# Batches: $B$; \# Epochs: $E$. 
		
		\Ensure Predicted Labels $\mathcal{Y}_U$, Parameters $\{\mathbf{W}^{l}\}_{l=0}^{L-1}$ and $f_\theta(\cdot)$.
		
		\State Initialize parameters $\{\mathbf{W}^{l}\}_{l=0}^{L-1}$, $f_\theta(\cdot)$, and $g_\gamma(\cdot)$.
		
		\For{$epoch$ $\in$ \{0,1,$\cdots$,$E-1$\}}
		    \For{$b$ $\in$ \{0,1,$\cdots$,$B-1$\}}
		        \State Sample a mini-batch of edges $\mathcal{E}_b$ from $\mathcal{E}$;
		        \State Compute losses $\mathcal{L}_{feat}$ and $\mathcal{L}_{label}$ by Eq.~(\ref{equ:7}) and Eq.~(\ref{equ:8});
		        \State Sum up $\mathcal{L}_{feat}$ and $\mathcal{L}_{label}$ as total loss $L_{total}$;
		        \State Update parameters by back-propagation of loss $L_{total}$.
		    \EndFor
		\EndFor
        \State Predict labels $\mathcal{Y}_U$ for those unlabeled nodes $\mathcal{V}_U$.
		\State \textbf{return} Predicted labels $\mathcal{Y}_U$, parameters $\{\mathbf{W}^{l}\}_{l=0}^{L-1}$ and $f_\theta(\cdot)$.
	\end{algorithmic}
\end{algorithm}

\subsection*{B. Rethink GNNs and GSDN from the perspective of information flow}
\begin{figure}[!htbp]
	\begin{center}
		\includegraphics[width=0.7\linewidth]{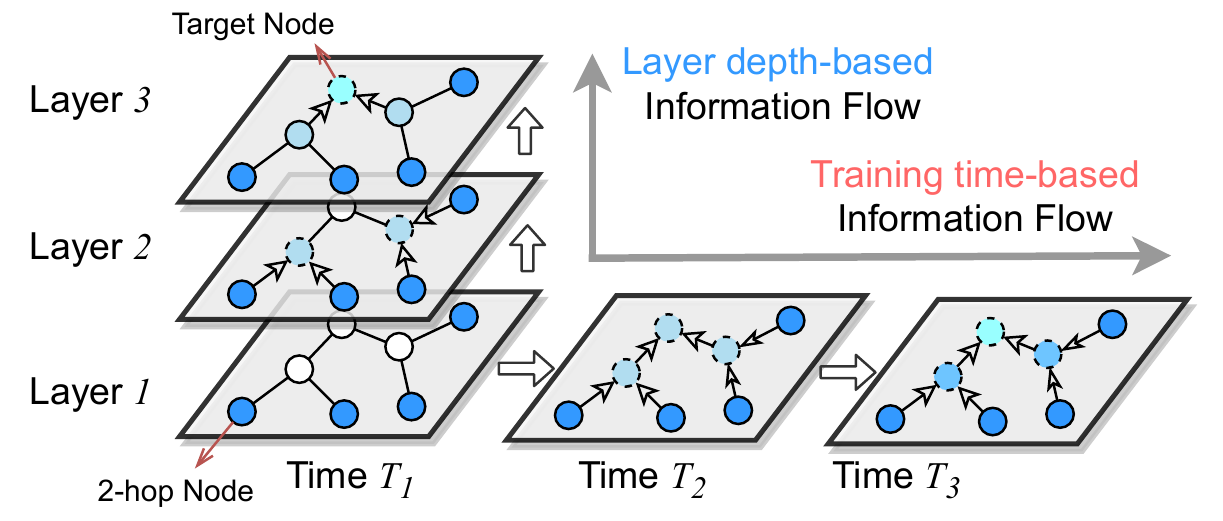}
	\end{center}
	\vspace{-1em}
	\caption{A comparison of two types of information flow, both of which work to capture information from long-range nodes. However, compared to the layer stacking of GNNs, GSDN works by cascaded neighborhood feature self-distillation, along the time dimension instead of stacking layers.}
	\label{fig:A1}
\end{figure}

\textbf{Discussion on information flow.}
We rethink the information flow of GNNs and GSDN from two perspectives, \textit{layer depth and training time}. (1) Considering message-passing-based GNNs, they rely on both training time and layer stacking to aggregate information, with \emph{the later playing a determining role}. Most existing GNNs enlarge the receptive field to multi-hop neighbors by stacking multiple GNN layers, such as graph convolution, to capture the long-distance dependency between nodes. In other words, the information of nodes that are multi-hop away flows to the target node along the unfolding path of the stacked GNN layers, which can be considered as a kind of information flow along the layer dimension. (2) In contrast, the layer depth of GSDN has no effect on the receptive field. Instead, as the training proceeds, information from the remote nodes gradually flows to the target node through the \emph{cascaded neighborhood feature self-distillation}, which can be viewed as information flow along the time dimension. (3) A comparison of these two types of information flow is shown in Fig.~\ref{fig:A1}, from which we find that at \textit{Layer 2}, the target node does not receive any information from 2-hop nodes. Instead, at \textit{Time 2}, not only does the target node receive information from 2-hop nodes, but also more information can be gradually distilled into the target node as training proceeds. In summary, while Eq.~(\ref{equ:4})(\ref{equ:6}) is defined on the 1-hop neighborhood, it can still propagate messages to multi-hops away, along the time dimension instead of stacking layers.

\subsection*{C. Dataset Statistics} 
\emph{Six} publicly available graph datasets are used to evaluate the proposed GSDN framework. An overview summary of the statistical characteristics of datasets is given in Tab.~\ref{tab:A1}. For the three small-scale datasets, namely Cora and Citeseer, we follow the data splitting strategy in \citep{kipf2016semi}. For the four large-scale datasets, namely Coauthor-CS, Coauthor-Physics, Amazon-Photo, and Amazon-Computers, we follow \citep{zhang2021graph,luo2021distilling} to randomly split the data into train/val/test sets, and each random seed corresponds to a different splitting.

\begin{table*}[!htbp]
\begin{center}
\caption{Statistical information of the datasets.}
\label{tab:A1}
\vspace{0.7em}
\resizebox{0.95\textwidth}{!}{
\begin{tabular}{lccccccc}

\toprule
\textbf{Dataset} & \texttt{Cora} & \texttt{Citeseer}  & \texttt{Amazon-Photo} & \texttt{Coauthor-CS} & \texttt{Coauthor-Phy} & \texttt{Amazon-Com} \\ \midrule
\textbf{$\#$ Nodes} & 2708 & 3327 & 7650 & 18333 & 34493 & 13752 \\
\textbf{$\#$ Edges} & 5278 & 4614 & 119081 & 81894 & 247962 & 245861 \\
\textbf{$\#$ Features} & 1433 & 3703 & 745 & 6805 & 8415 & 767 \\
\textbf{$\#$ Classes} & 7 & 6 & 8 & 15 & 5 & 10 \\
\textbf{Label Rate} & 5.2\% & 3.6\% & 2.1\% & 1.6\% & 0.3\% & 1.5\% & \\ \bottomrule

\end{tabular}}
\end{center}
\end{table*}

\subsection*{D. Parameter Sensitivity Analysis (Q5)}
\begin{figure}[!htbp]
	\begin{center}
		\includegraphics[width=0.40\linewidth]{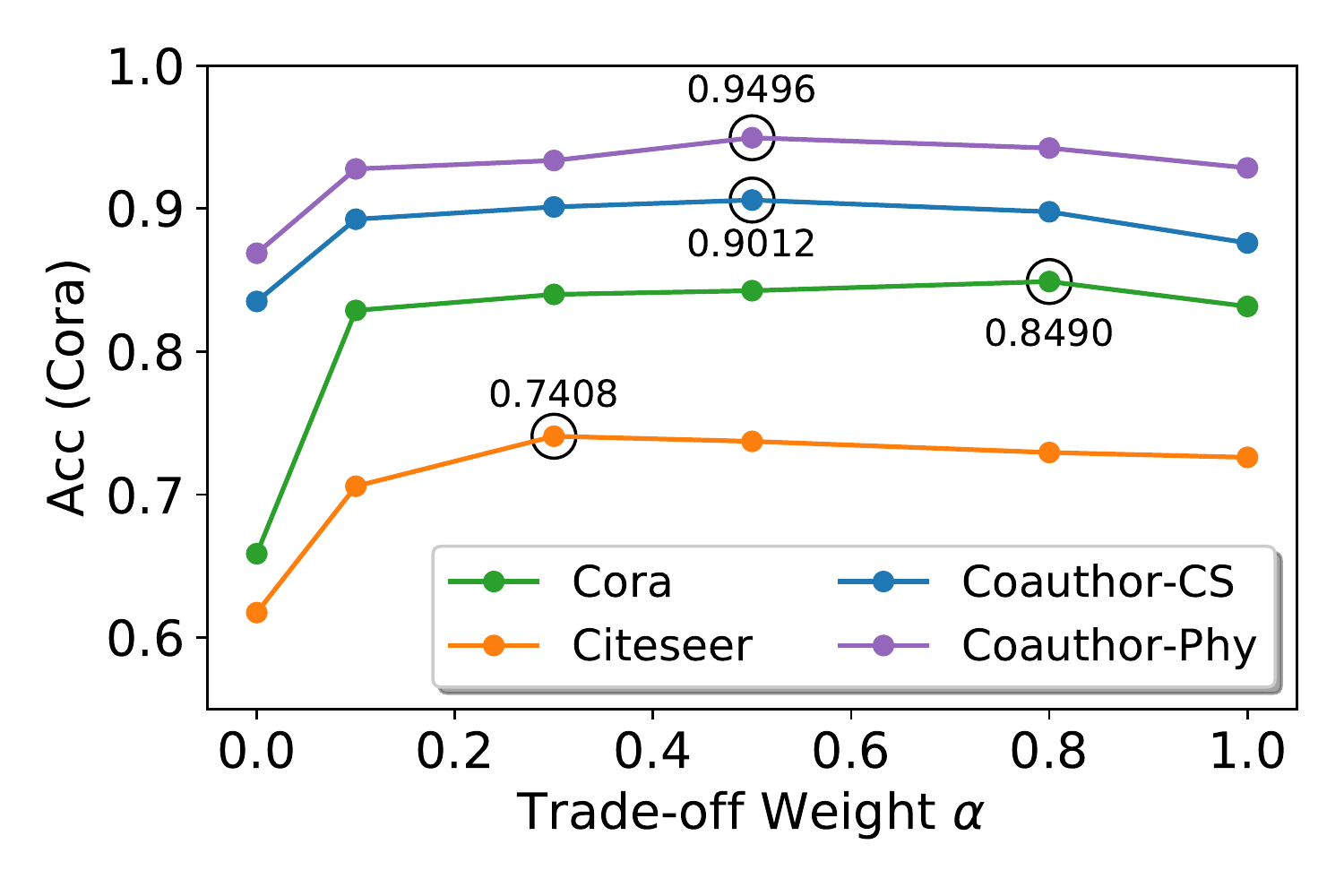}
		\includegraphics[width=0.39\linewidth]{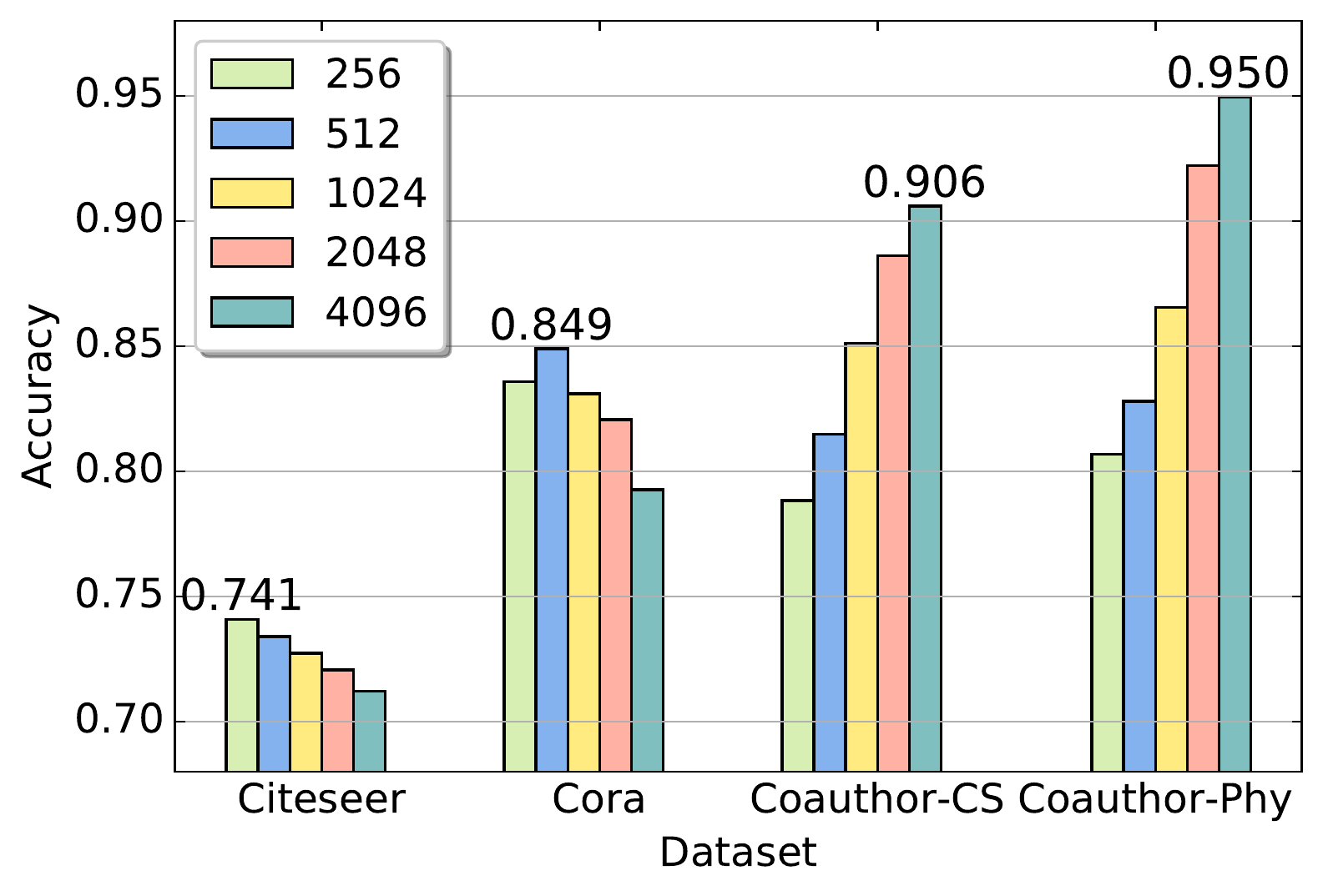}
	\end{center}
	\vspace{-1em}
	\caption{Parameter sensitivity analysis on the trade-off weight $\lambda$ and batch size $B$ on four datasets.}
	\label{fig:A2}
\end{figure}

To answer \textbf{Q5}, we evaluate the parameter sensitivity w.r.t two key hyperparameters: trade-off weight $\lambda\in\{0.0, 0.1, 0.3, 0.5, 0.8, 1.0\}$ and batch size $B\in\{256, 512, 1024, 2048, 4096\}$ in Fig.~\ref{fig:A2}, from which we can observe that (1) batch size $B$ is a dataset-specific hyperparameter. For simple graphs with few nodes and edges, such as Cora, a small batch size, $B=256$, can yield fairly good performance. However, for large-scale graphs with more nodes and edges, such as Coauthor-Phy, the model performance usually improves with the increase of batch size $B$. (2) When $\lambda$ is set to 0, i.e., the neighborhood feature-level self-distillation is completely removed, the performance of GSDN degrades to be close to that of MLPs. In contrast, when $\lambda$ takes a non-zero value, the performance of GSDN improves as $\lambda$ increases. However, when $\lambda$ becomes too large, it weakens the benefit of label information, yielding lower performance gains. In practice, we can usually determine $B$ and $\lambda$ by selecting the model with the highest accuracy on the validation set through the grid search.

\subsection*{E. More results on Performance with Noisy Labels.} 
\begin{figure}[!htbp]
	\begin{center}
		\includegraphics[width=0.38\linewidth]{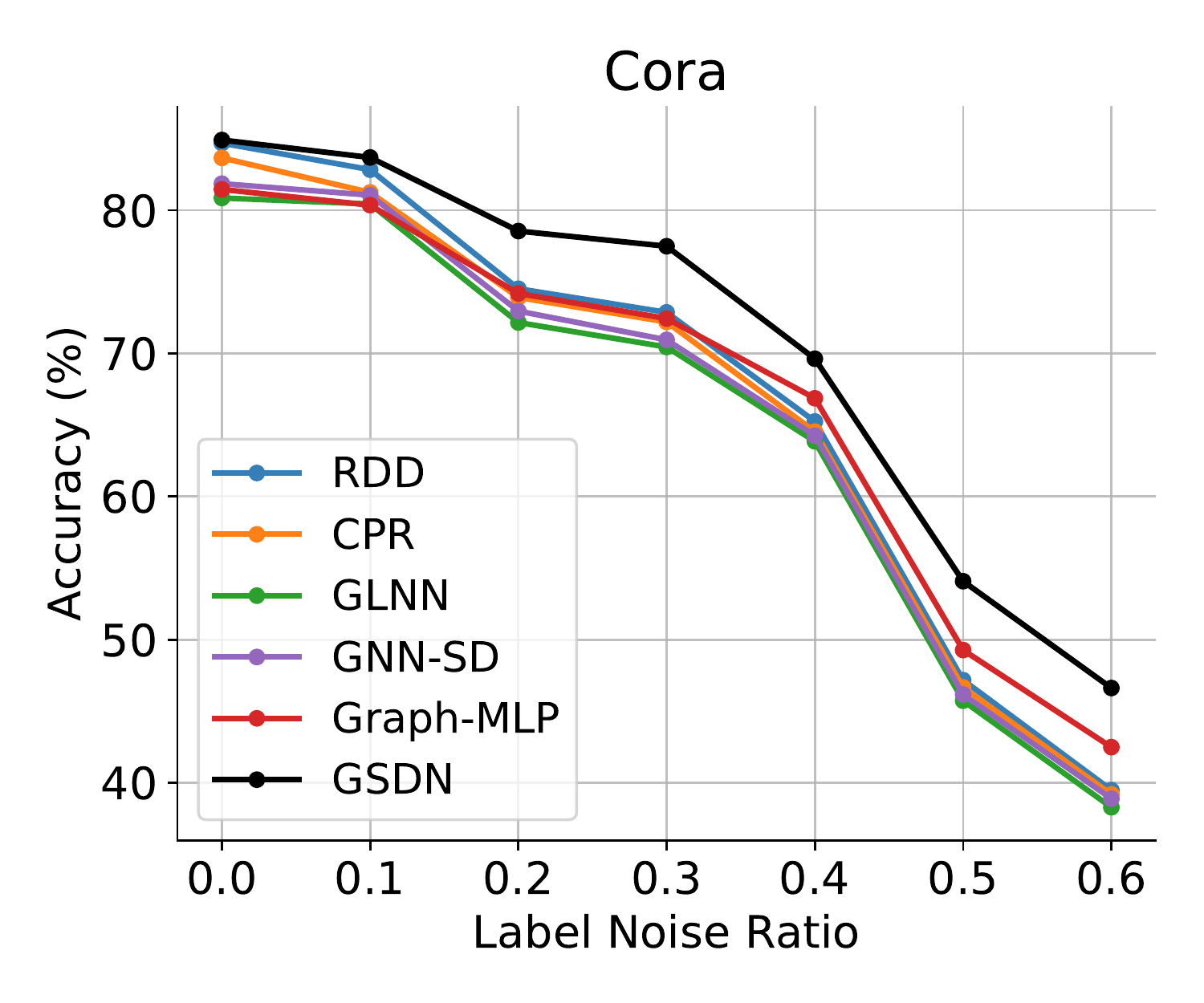}
		\includegraphics[width=0.38\linewidth]{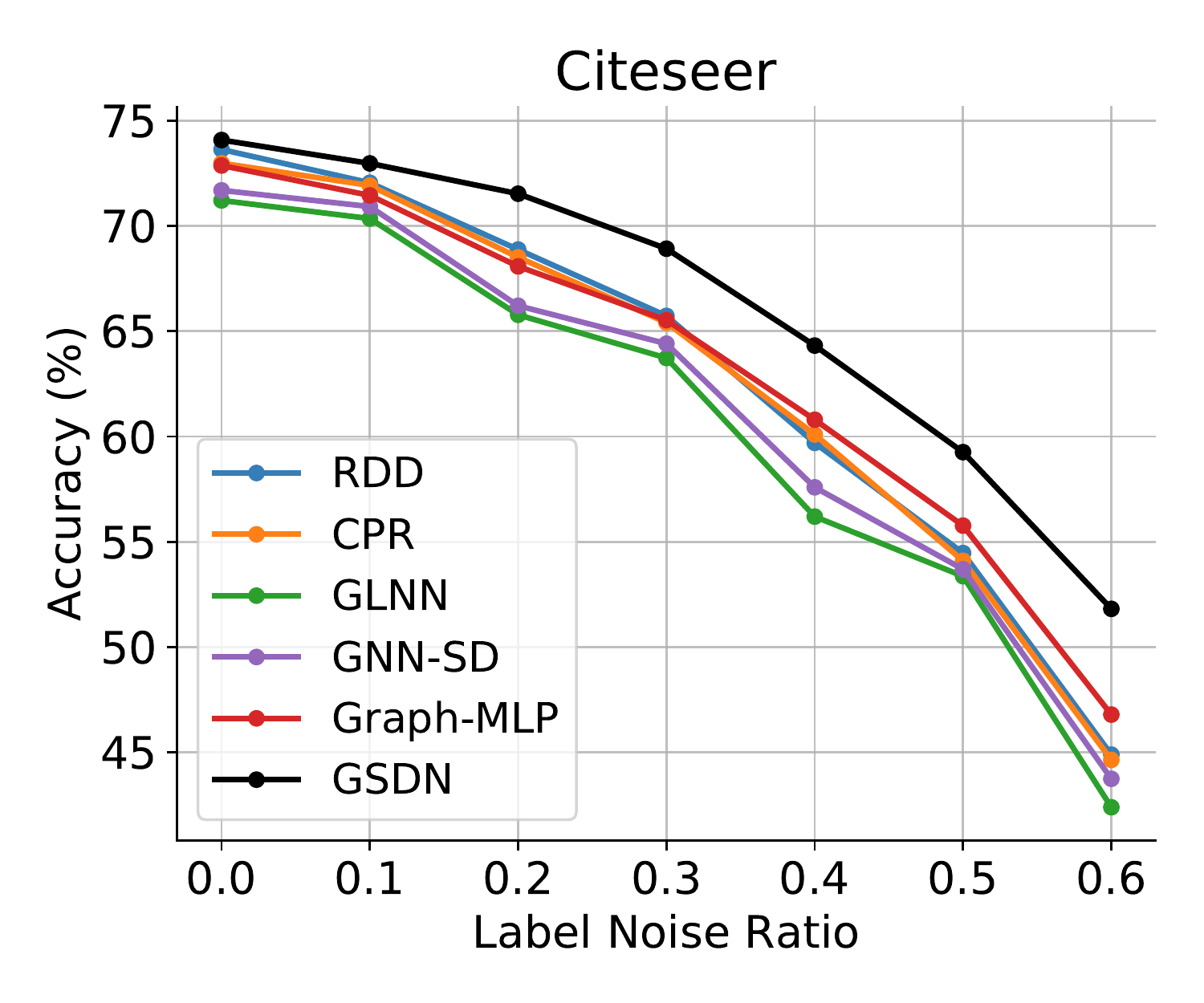}
	\end{center}
	\vspace{-1em}
	\caption{Accuracy (\%) under different label noise ratios on the Cora and Citeseer datasets.}
	\vspace{-1em}
	\label{fig:A3}
\end{figure}

The performance comparison with the graph distillation methods and MLP-based models under noisy labels is reported in Fig.~\ref{fig:A3}. As can be seen, the accuracy of GSDN drops more slowly than other baselines as the noise ratio $r$ increases, and GSDN is more robust than other models under various label noise ratios, especially under extremely high noise ratios. Besides, we find that while graph distillation methods perform well on clean data, as shown in Table.~\ref{tab:1}, their performance gains are reduced when the label noise ratio $r$ increases. In contrast, MLP-based models, both Graph-MLP and GSDN, show great advantages over GNN-based models under extremely high noise ratios.

\subsection*{F. Model accuracy \textit{vs.} Inference Time.} 
To provide a high-level picture of the trade-off between model accuracy and inference time, we show the performance of various methods on the Cora dataset in Fig.~\ref{fig:3}, where we can divide all methods into the following four categories based on their \textbf{inference accuracy and inference speed}: (1) \textit{high-accuracy and high-latency}. Existing general GNN models and most GNN distillation models can achieve high classification accuracy, but still suffer from inference latency caused by data dependency. (2) \textit{low-accuracy and low-latency}. The pure MLP-based models and GLNN enjoy the benefit of low inference latency, but their classification accuracy cannot match that of the current state-of-the-art GNNs. (3) \textit{low-accuracy and high-latency}. The inference acceleration methods, such as pruning, quantization, and sampling, only have small gains in the inference speed but greatly hurt the inference accuracy, as they have not fully addressed the neighborhood fetching problem. (4) \textit{high-accuracy and low-latency}. The proposed GSDN framework matches the state-of-the-art GNNs in terms of classification accuracy and also is comparable to MLPs in the inference speed.

\subsection*{I. Training Time}
The training time complexity of the proposed GSDN framework is $\mathcal{O}(|\mathcal{V}|dF+|\mathcal{E}|F)$, which is linear with respect to the number of nodes $|\mathcal{V}|$ and edges $|\mathcal{E}|$, and is in the same order of magnitude as GCN. The training time ($s$) averaged over 30 sets of runs on four datasets is reported in Table.~\ref{tab:3} with the time multiple w.r.t the vanilla GCN marked as ${\color[rgb]{0.4, 0.71, 0.376}green}$, where all methods use $L=2$ layers and hidden dimension $F=16$.  From Table.~\ref{tab:A2}, we have the following observations: \textit{(1)} Common inference acceleration methods, including pruning, quantization and neighborhood sampling, not only help speed up GNN inference, but also work for GNN training. \textit{(2)} While GLNN, Graph-MLP and GSDN have great speed advantages in the inference stage, they essentially shift considerable work from the latency-sensitive inference stage, where time reduction in milliseconds makes a huge difference, to the less latency-insensitive training stage, where time cost in hours or days is often tolerable. This explains why these methods require more training time compared to the vanilla GCN. \textit{(3)} Also as a MLP-based model, \textbf{the training time of GSDN is almost half of that of Graph-MLP}, mainly because Graph-MLP performs contrasting within higher-order neighborhoods, while GSDN only considers first-order neighbors, which greatly reduces the computational burden in training.

\begin{table}[!htbp]
\begin{center}
\caption{Training time ($s$), where the time multiple w.r.t the vanilla GCN is marked as ${\color[rgb]{0.4, 0.71, 0.376}green}$.}
\label{tab:A2}
\vspace{0.5em}
\resizebox{\columnwidth}{!}{
\begin{tabular}{lccccccccc}

\toprule
\textbf{Method}       & GCN   & APPNP & DAGNN & P-GCN & Q-GCN & NS-GCN & GLNN                                                & Graph-MLP                                            & GSDN (ours)                                          \\ \midrule
\textbf{Cora}         & 12.8  & 25.2  & 33.3  & 8.1  & 7.5  & 6.3    & 20.4 (${\color[rgb]{0.4, 0.71, 0.376}1.59\times}$)  & 56.3 (${\color[rgb]{0.4, 0.71, 0.376}4.40\times}$)   & 30.1 (${\color[rgb]{0.4, 0.71, 0.376}2.35\times}$)   \\
\textbf{Citeseer}     & 23.5  & 34.7  & 42.4  & 15.2  & 13.7  & 12.6   & 31.8 (${\color[rgb]{0.4, 0.71, 0.376}1.35\times}$)  & 71.8 (${\color[rgb]{0.4, 0.71, 0.376}3.06\times}$)   & 43.6 (${\color[rgb]{0.4, 0.71, 0.376}1.86\times}$)   \\
\textbf{Coauthor-CS}  & 219.3 & 355.6 & 560.3 & 146.2 & 127.7 & 116.9  & 286.9 (${\color[rgb]{0.4, 0.71, 0.376}1.31\times}$) & 1029.6 (${\color[rgb]{0.4, 0.71, 0.376}4.69\times}$) & 648.64 (${\color[rgb]{0.4, 0.71, 0.376}2.95\times}$) \\
\textbf{Coauthor-Phy} & 507.2 & 799.7 & 682.9 & 334.9 & 322.8 & 284.4  & 656.2 (${\color[rgb]{0.4, 0.71, 0.376}1.29\times}$) & 1852.1 (${\color[rgb]{0.4, 0.71, 0.376}3.65\times}$) & 1076.8 (${\color[rgb]{0.4, 0.71, 0.376}2.12\times}$) \\ \bottomrule

\end{tabular}}
\end{center}
\end{table}

\clearpage

\end{document}